\documentclass[sigconf]{acmart}
\AtBeginDocument{%
  \providecommand\BibTeX{{%
    \normalfont B\kern-0.5em{\scshape i\kern-0.25em b}\kern-0.8em\TeX}}}

\copyrightyear{2021}
\acmYear{2021}
\setcopyright{acmcopyright}
\acmConference[KDD '21] {Proceedings of the 27th ACM SIGKDD Conference on Knowledge Discovery and Data Mining}{August 14--18, 2021}{Virtual Event, Singapore.}
\acmBooktitle{Proceedings of the 27th ACM SIGKDD Conference on Knowledge Discovery and Data Mining (KDD '21), August 14--18, 2021, Virtual Event, Singapore}
\acmPrice{15.00}
\acmISBN{978-1-4503-8332-5/21/08}
\acmDOI{10.1145/3447548.3467397}
\usepackage{xspace}
\usepackage{enumitem}
\usepackage[ruled,vlined]{algorithm2e}
\usepackage[font=small, skip=4pt]{caption}

\usepackage{booktabs}
\usepackage{multirow}
\usepackage{subfigure}
\usepackage{nicefrac}
\usepackage{url}
\usepackage{makecell}

\newcommand{\our}{\mbox{UCPhrase}\xspace}

\newcommand{\ie}{\mbox{i.e.}\xspace}
\newcommand{\eg}{\mbox{e.g.}\xspace}

\newcommand{\vpar}[1]{\vspace{.2em}\noindent{\textbf{#1}}}

\newcommand{\concept}[1]{\textbf{\emph{#1}}}
\newcommand{\example}[1]{``\emph{#1}''}

\newcommand{\Rom}[1]{\uppercase\expandafter{\romannumeral #1}}

\newcommand{\smtx}[1]{\scriptsize{#1}}
\setlist{nosep,after=\vspace{\parskip}}

\newcommand{\tightmath}[1]{\everymath{\medmuskip=1.5mu minus 1.5mu\thickmuskip=2mu minus 2mu}$#1$\everymath{\medmuskip=2mu minus 2mu\thickmuskip=4mu minus 4mu}}
\usepackage[utf8]{inputenc}
\usepackage{ascii}
\usepackage{newunicodechar}

\usepackage{amsmath}
\DeclareMathOperator*{\argmin}{argmin}

\definecolor{gred}{RGB}{219,68,55}
\definecolor{gblue}{RGB}{66,133,244}
\definecolor{gyellow}{RGB}{244,180,0}
\definecolor{ggreen}{RGB}{15,157,88}
\definecolor{ggrey}{RGB}{115,115,115}
\newcommand{\algcomment}[1]{\textcolor{gblue}{{#1}}} 

\newcommand{\catag}[1]{\textcolor{ggreen}{[\textbf{#1}]}}
\newcommand{\cetag}[1]{\textcolor{gblue}{[\textbf{#1}]}}
\newcommand{\watag}[1]{\textcolor{gyellow}{[\textbf{#1}]}}
\newcommand{\wetag}[1]{\textcolor{gred}{[\textbf{#1}]}}

\newcommand{\myfootnote}[2]{\footnote{\scriptsize{#1} \href{#2}{\scriptsize{#2}}}}

\begin{document}

\widowpenalty=10

\title{\our: Unsupervised Context-aware Quality Phrase Tagging}

\author{Xiaotao Gu$^{1*}$, Zihan Wang$^{2*}$, Zhenyu Bi$^{2}$, Yu Meng$^1$, Liyuan Liu$^1$, Jiawei Han$^1$, Jingbo Shang$^2$}
\affiliation{
\institution{$^1$University of Illinois at Urbana-Champaign $\quad$ \{xiaotao2, yumeng5, ll2, hanj\}@illinois.edu}
\institution{ $^2$University of California San Diego $\quad$ \{ziw224, z1bi, jshang\}@ucsd.edu}
\country{}
}
\renewcommand{\shortauthors}{Xiaotao Gu and Zihan Wang, et al.}

\thanks{$^*$Equal Contribution.}
\begin{abstract}
Identifying and understanding \emph{quality phrases} from context is a fundamental task in text mining.  
The most challenging part of this task arguably lies in uncommon, emerging, and domain-specific phrases. 
The infrequent nature of these phrases significantly hurts the performance of phrase mining methods that rely on sufficient phrase occurrences in the input corpus.
Context-aware tagging models, though not restricted by frequency, heavily rely on domain experts for either massive sentence-level gold labels or handcrafted gazetteers.
In this work, we propose \our, a novel unsupervised context-aware quality phrase tagger.
Specifically, we induce high-quality phrase spans as silver labels from consistently co-occurring
word sequences within each document.
Compared with typical context-agnostic distant supervision based on existing knowledge bases (KBs), our silver labels root deeply in the input domain and context, thus having unique advantages in preserving
contextual completeness 
and capturing emerging, out-of-KB phrases.
Training a conventional neural tagger based on silver labels usually faces the risk of overfitting phrase surface names.
Alternatively, we observe that the contextualized attention maps generated from a Transformer-based neural language model effectively reveal the connections between words in a surface-agnostic way.
Therefore, we pair such attention maps with the silver labels to train a lightweight span prediction model, which can be applied to new input to recognize (unseen) quality phrases regardless of their surface names or frequency. 
Thorough experiments on various tasks and datasets, including corpus-level phrase ranking, document-level keyphrase extraction, and sentence-level phrase tagging, demonstrate the superiority of our design over state-of-the-art pre-trained, unsupervised, and distantly supervised methods.

\end{abstract}


\maketitle
\begin{figure*}
    \centering
    \includegraphics[width=1\linewidth]{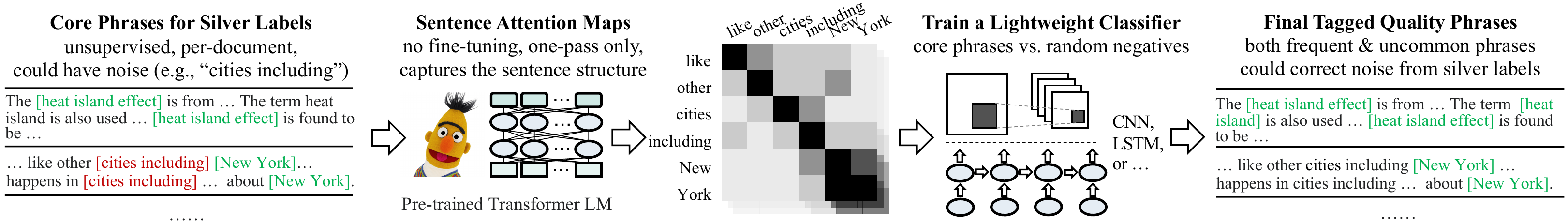}
    \vspace{-1.5em}
    \caption{
    An overview of our \our: unsupervised context-aware quality phrase tagging.
    }%
    \label{fig:pipeline}
    \vspace{-2.5mm}
\end{figure*}

\section{Introduction}

Quality phrases refer to informative multi-word sequences that ``\emph{appear consecutively in the text, forming a complete semantic unit in certain contexts or the given document}''~\cite{finch2016linguistic}.
Identifying and understanding quality phrases from context is a fundamental task in text mining.
Automated quality phrase tagging serves as a cornerstone in a broad spectrum of downstream applications, including but not limited to 
entity recognition~\cite{shang2018learning},
text classification~\cite{altinel2018semantic},
and information retrieval~\cite{croft1991use}.

The most challenging open problem in this task is how to recognize uncommon, emerging phrases, especially in specific domains. 
These phrases are essential in the sense of their significant semantic meanings and the large volume---following a typical Zipfian distribution, uncommon phrases can add up to a significant portion of quality phrases \cite{williams2015zipf}.
Moreover, emerging phrases are critical in understanding domain-specific documents, such as scientific papers, since new terminologies often come along with transformative innovations.
However, mining such sparse long-tail phrases is nontrivial, since a frequency threshold has long ruled them out in traditional phrase mining methods~\cite{deane2005nonparametric,li2017efficiently,el2014scalable,liu2015mining,shang2018automated} due to the lack of reliable frequency-related corpus-level signals (e.g., the mutual information of its sub-ngrams). 
For instance, AutoPhrase~\cite{shang2018automated} only recognizes phrases with at least 10 occurrences by default.

For infrequent phrases, the tagging process largely relies on local context.
Recent advances in neural language models have unleashed the power of sentence-level contextualized features in building chunking- and tagging-based models~\cite{manning2014stanford,wang2020mining}.
These context-aware models can even recognize unseen phrases from new input texts, thus being no longer restricted by frequency.
However, training a domain-specific tagger of reasonably high quality requires expensive, hard-to-scale effort from domain experts for massive sentence-level gold labels or handcrafted gazetteers.

In this work, we propose \our, a novel unsupervised context-aware quality phrase tagger.
It first induces high-quality silver labels directly from the corpus under the unsupervised setting, and then trains a tailored Transformer-based neural model that can recognize quality phrases in new sentences. 
Figure~\ref{fig:pipeline} presents an overview of \our. 
The two major steps are detailed as follows.

By imitating the reading process of humans, we derive supervision directly from the input corpus.
Given a document, human readers can quickly recognize new phrases or terminologies from the consistently used word sequences \emph{within the document}.
The ``document'' here refers to a collection of sentences centered on the same topic, such as sentences from an abstract of a scientific paper and tweets mentioning the same hashtag. 
Inspired by this observation, we propose to extract \concept{core phrases}
from each document, which are maximal contiguous word sequences that appear in the document more than once.
The ``maximal'' here means that if one expands this word sequence further towards the left or right, its frequency within this document will drop.
To avoid uninformative phrases (\eg, \example{of a}), we conduct simple filtering of stopwords before finalizing the silver labels. 
Note that our proposed silver label generation follows a per-document manner.
Therefore, compared with typical context-agnostic distant supervision based on existing knowledge bases or dictionaries~\cite{shang2018automated, shang2018learning, wang2020mining}, our supervision roots deeply in the input domain and context, thus having unique advantages in preserving contextual completeness of matched spans and capturing much more emerging phrases.

We further design a tailored neural tagger to fit our silver labels better. 
Training a conventional neural tagger based on silver labels usually faces a high risk of overfitting the observed labels~\cite{liang2020bond}.
With access to the word-identifiable embedding features, it is easy for the model to achieve nearly zero training error by rigidly memorizing the surface names of training labels.
Alternatively, we find that the contextualized attention distributions generated from a Transformer-based neural language model could capture the connections between words in a surface-agnostic way~\cite{kim2019pre}.
Intuitively, the attention maps of quality phrases should reveal distinct patterns from ordinary word spans.
Moreover, attention-based features block the direct access to the surface names of training labels, and force the model to learn about more general context patterns.
Therefore, we pair such surface-agnostic features based on attention maps with the silver labels to train a neural tagging model, which can be applied to new input to recognize (unseen) quality phrases.
Specifically, given an unlabeled sentence of $N$ words, we first encode the sentence with a pre-trained Transformer-based language model and obtain the attention maps as features.
The $N \times N$ matrices from different Transformer layers and attention heads can be viewed as images to be classified with multiple channels.
A lightweight CNN-based classifier is then trained to distinguish quality phrases from randomly sampled negative spans.

Thorough experiments on various tasks and datasets, including corpus-level phrase ranking, document-level keyphrase extraction, and sentence-level phrase tagging, demonstrate the superiority of our design over state-of-the-art unsupervised, distantly supervised methods, and pre-trained off-the-shelf tagging models.
It is noteworthy that our trained model is robust to the noise in the core phrases---our case studies in Section~\ref{sec:case_study} show that the model can identify inferior training labels by assigning extremely low scores.

Efficiency wise, thanks to the rich semantic and syntactic knowledge in the pre-trained language model, we can simply use the generated attention maps as informative features without fine-tuning the language model.
Hence we only need to update the lightweight classification model during training, making the training process as fast as one inference pass of the language model through the corpus with limited resource consumption.

\smallskip\noindent
To the best of our knowledge, \our is the first unsupervised context-aware quality phrase tagger.
It enjoys the rich knowledge from the pre-trained neural language models.
The learned phrase tagger works efficiently and effectively without reliance on human annotations, existing knowledge bases, or phrase dictionaries.
We summarize our key contributions as follows:
\begin{itemize}[leftmargin=*,nosep]
\item We propose to mine silver labels that root deeply in the input domain and context by recognizing core phrases, \ie, maximal word sequences that occur consistently in a per-document manner. 
\item We propose to replace the conventional contextualized word representations with surface-agnostic attention maps generated by pre-trained Transformer-based language models to alleviate the risk of overfitting silver labels.
\item We conduct extensive experiments, ablation studies, and case studies to compare \our with state-of-the-art unsupervised, distantly supervised methods, and pre-trained off-the-shelf tagging models. The results verify the superiority of our method
\footnote{Code and data: \href{https://github.com/xgeric/UCPhrase-reproduce}{\color{blue}https://github.com/xgeric/UCPhrase-exp.}}. 
\end{itemize}

\section{Problem Definition}

Given a sequence of words $[w_1, \ldots, w_N]$, a quality phrase is a contiguous span of words $[w_i, \ldots, w_{i+k}]$ that form a complete and informative semantic unit in context.
Though some studies also view unigrams as potential phrases, in this work, we focus on multi-word phrases ($k > 0$), which are more informative, yet more challenging to get due to both diversity and sparsity.

To effectively capture phrases with potential overlaps, \eg, \example{information extraction} in \example{information extraction systems}, 
we adopt the span prediction framework, where each possible span in the sentence is assigned a binary label.
To avoid a quadratic growth of the size of candidate spans, we follow previous work \cite{liu2015mining, shang2018automated} to set a maximum span length $K$.
We also explore alternative classifiers based on the sequence labeling framework in Section \ref{sec:method}.

\begin{figure*}[t]
    \centering
    \subfigure[Example Silver Labels]{
        \label{fig:supervision_example}
        \centering
        \includegraphics[width=0.42\linewidth]{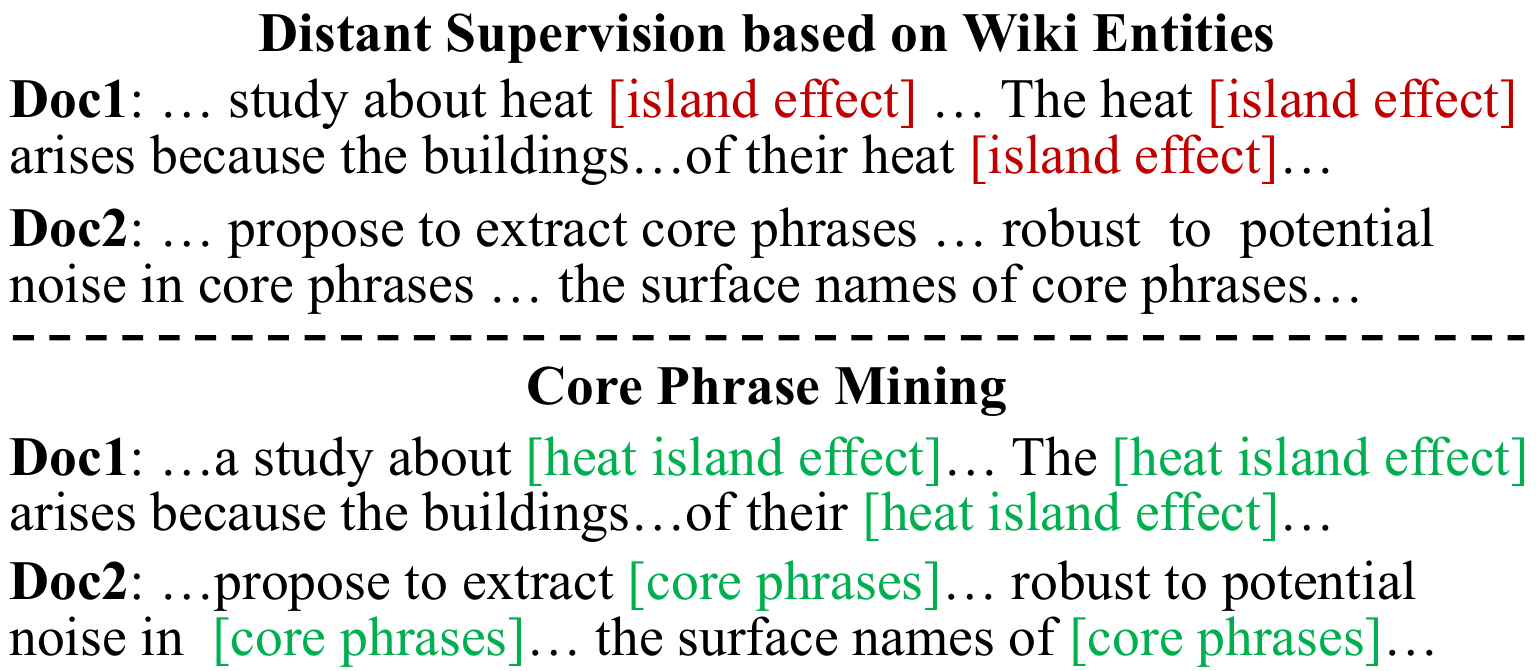}
    }
    \subfigure[Occurrence Distribution]{
        \label{fig:supervision_occ}
        \centering
        \includegraphics[width=0.27\linewidth]{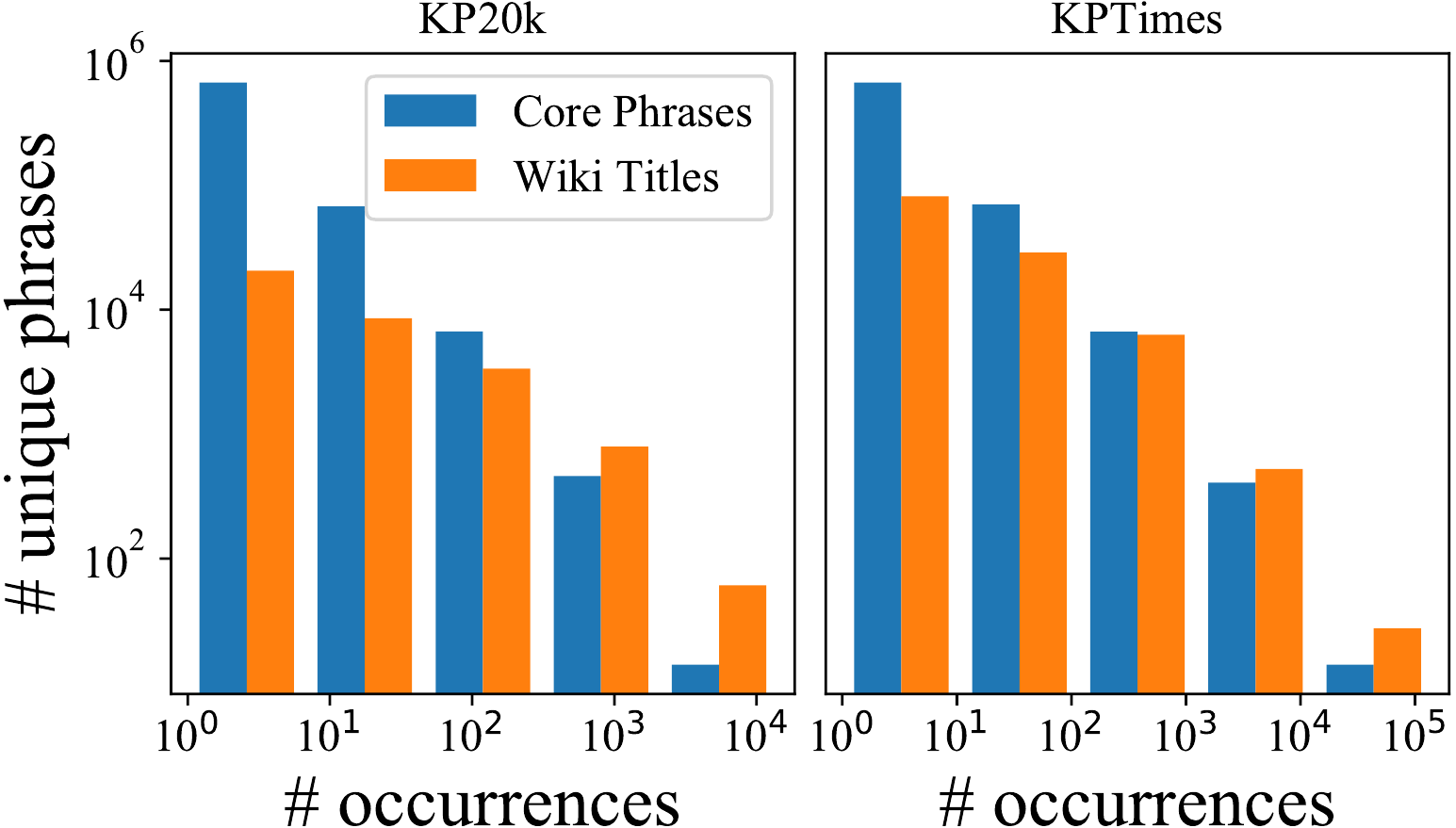}
    }
    \subfigure[Length Distribution]{
        \label{fig:supervision_len}
        \centering
        \includegraphics[width=0.27\linewidth]{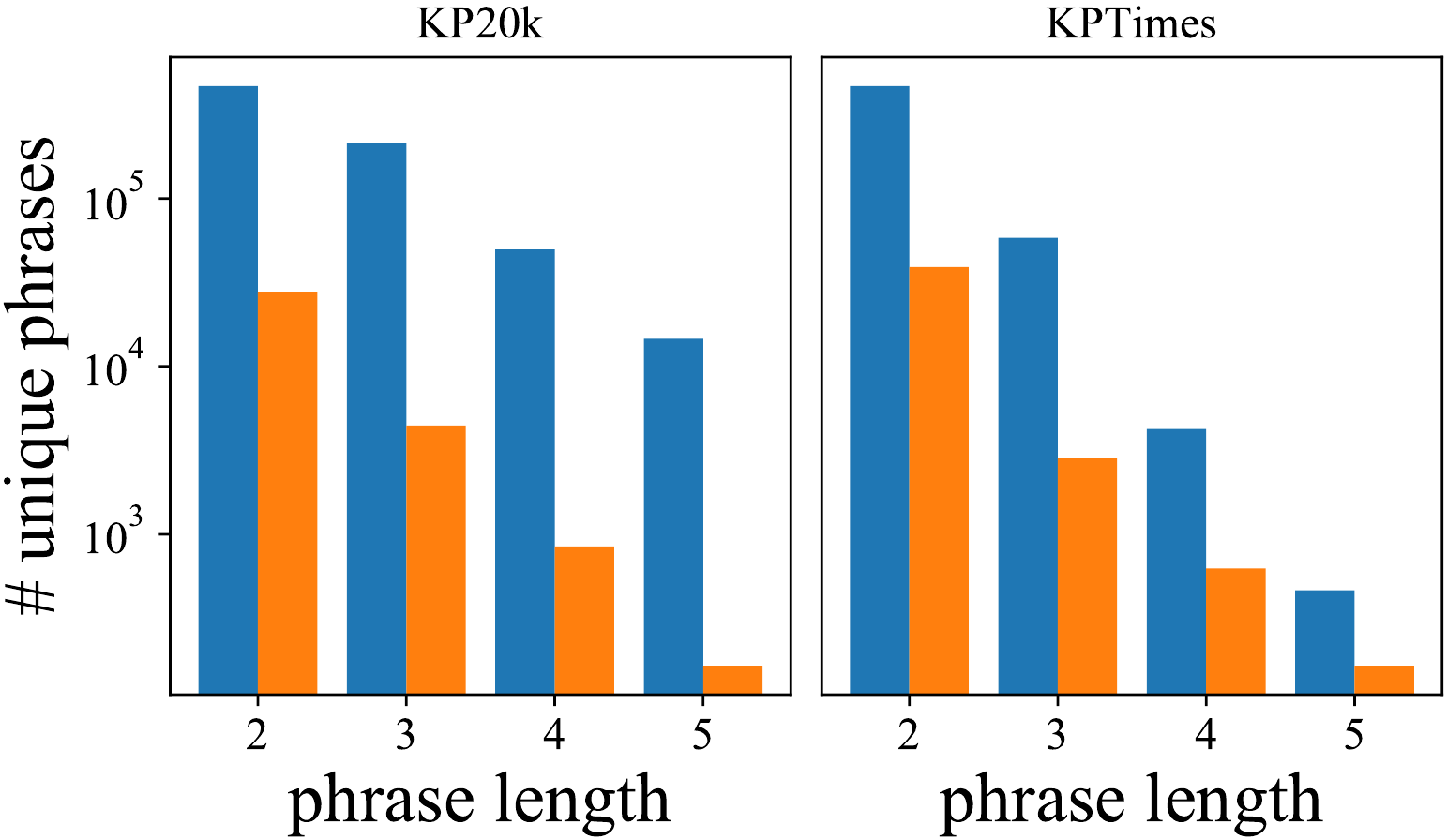}
    }
    \vspace{-2mm}
    \caption{
    Comparing core phrases with context-agnostic distant supervision.
    (a) An illustrative example with context.
    Our core phrases preserve better contextual completeness and discover emerging new concepts introduced in the document.
    (b) Distributions of the generated silver labels with their occurrences in the corpus.
    The X-axis represents bins of phrase occurrences in the corpus. The Y-axis (exponential) represents the number of unique phrases in each bin. 
    (c) Distributions of the generated silver labels with their lengths (\# of words).
    }
    \label{fig:supervision}
    \vspace{-2.2mm}
\end{figure*}

\section{\our: Methodology}
\label{sec:method}

Figure \ref{fig:pipeline} presents an overview of \our.
As an unsupervised method, \our first mines core phrases directly from each document as silver labels and extracts surface-agnostic attention features with a pre-trained language model.
A lightweight classifier is then trained with silver labels and randomly sampled negative labels.
Algorithm \ref{alg:main} shows the detailed training process.

\begin{algorithm}[t]
\small
	\KwIn{A corpus of $M$ unlabeled documents, $\{d_m\}^M_{m=1}$; \\a pre-trained language model (LM); the attention-based lightweight classification model $f(\cdot; \theta)$ to be trained.}
	
	\algcomment{// Generate silver labels as supervision (Sec.~\ref{sec:labelgeneration})} \\ 
	\For{each document $d_m$} {
	    $\mathcal{P}^+_m = MineCorePhrases(d_m)$. \\
	   Randomly sample remaining spans as negative set $\mathcal{P}^-_m$.
	}
	
	\algcomment{// Generate surface-agnostic features using LM (Sec.~\ref{sec:attentionfeature})} \\ 
	\vspace{-.2em}
	\For{each labeled span $p_i \in \mathcal{P} = \bigcup\limits_{M} \{\mathcal{P}_m^-, \mathcal{P}_m^+ \}$} {
	    $\mathbf{X}_{p_i} = ExtractAttentionFeature(\text{LM}, p_i)$
	}
	$\mathcal{D}_{train}, ~~~ \mathcal{D}_{valid} = split(\mathbf{X}_{\mathcal{P}}, \mathcal{P})$.
	
	\algcomment{// Train a classifier based on features \& labels (Sec.~\ref{sec:classifier})} \\
	\Repeat{F$_1$ score on $\mathcal{D}_{valid}$ drops}{
	  Sample a minibatch $\{\mathbf{X}_{\mathcal{P}_b}, \mathcal{P}_b\}$ from $\mathcal{D}_{train}$.\\
	  Update model parameters with loss $\ell(f(\mathbf{X}_{\mathcal{P}_b}; \theta), \mathcal{P}_b)$.
	}
	\KwOut{The trained classification model $f$.}
	\caption{\our: unsupervised model training}
	\label{alg:main}
\end{algorithm}

\subsection{Silver Label Generation}
\label{sec:labelgeneration}

As the first step, we seek to collect high-quality phrases in the input corpus following an unsupervised way, which will be our silver labels for the tagging model training.
A common practice for automated label fetching is to conduct a context-agnostic matching between the corpus and a given quality phrase list, either mined with unsupervised models or collected from an existing knowledge base (KB).
Such methods, as we show later, can suffer from incomplete labels due to the negligence of context.

On the contrary, based on the definition of phrases, we look for consistently used word sequences in context.
We propose to treat documents as context and collect high-quality phrases directly from each document.
The ``document'' here refers to a collection of sentences centered on the same topic, such as sentences from an abstract of a scientific paper and tweets mentioning the same hashtag. 
This way, we expect to preserve better contextual completeness that reflects the original writing intention.

We view a document $d_m$ as a contiguous word sequence [$w_1$, $\ldots$, $w_N$] and then mine max contiguous sequential patterns.
A valid pattern here is a word span $[w_i, \ldots, w_j]$ that appear more than once in the input sequence.
One can easily adjust the frequency threshold to balance the quality and quantity of valid patterns.
In this work, we simply use the minimum requirement of two occurrences without further tuning, and find it works well for both short documents like paper abstracts and long documents like news reports. 
To preserve completeness, we only leave max patterns that are not sub-patterns of any other valid patterns.
Uninformative patterns like \example{of a} are removed with a stopword list widely used by previous work~\cite{liu2015mining,shang2018automated}.
We treat the remaining max patterns as core phrases of document $d_m$, and add them to the positive training samples $\mathcal{P}_m^+$.
An equal number of negative samples are randomly drawn from the remaining spans in $d_m$, denoted as $\mathcal{P}_m^-$.

Figure~\ref{fig:supervision} compares the silver labels generated by our core phrases with those by distant supervision, which follows a context-agnostic string matching from the Wikipedia entities.
From the real example in Figure~\ref{fig:supervision_example}, \example{heat island effect} is not a Wikipedia entity, but \example{island effect} is one.
Distant supervision hence generates a flawed label by partially matching the real phrase.
Similar examples are quite common especially when it comes to compound phrases, like \example{biomedical data mining}.
Distant supervision would tend to favor those popular phrases and generate incomplete matches in context.
On the contrary, our core phrase mining can generate labels with better contextual completeness. 
Core phrase mining can also dynamically capture concepts or expressions newly introduced in each document, such as the \example{core phrases} in this paper. 
 Figure~\ref{fig:supervision_occ} confirms this by showing the distribution of unique phrases of the two types of silver labels, mined from the KP20k CS publication corpus and the KPTimes news corpus, with respect to their frequency.
In particular, core phrase mining discovers more unique phrases with less than $10$ occurrences in the corpus (30x on KP20k, 9x on KPTimes).
As Figure~\ref{fig:supervision_len} demonstrates, core phrases outnumber matched Wiki titles on all length ranges.
Overall, core phrase mining discovers much more unique phrases than distant supervision (20x on KP20k, 6x on KPTimes).

Of course, there also inevitably exist noises in mined core phrases due to random word combinations consistently used in some documents, \eg, \example{countries including}.
Fortunately, since we collected core phrases from each document independently, such noisy labels will not spread and be amplified to the entire corpus.
In fact, among the tagged core phrases randomly sampled from two datasets, the overall proportion of high-quality labels is over $90\%$.
The large volume of reasonably high-quality silver labels provides a robust foundation for us to train a span classifier that learns about general context patterns to distinguish noisy spans.
As Section~\ref{sec:case_study} shows, the final classifier can assign extremely low scores to false-positive phrases in training labels.

In summary, document-level core phrase mining provides a simple and effective way to automatically fetch abundant context-aware silver labels of reasonably good quality without relying on external KBs.
In ablation studies (Section~\ref{sec:ablation-supervision}) we show that models trained with such free silver labels can outperform the same models trained with distant supervision.

\begin{figure}[t]
    \centering
    \includegraphics[width=.75\linewidth]{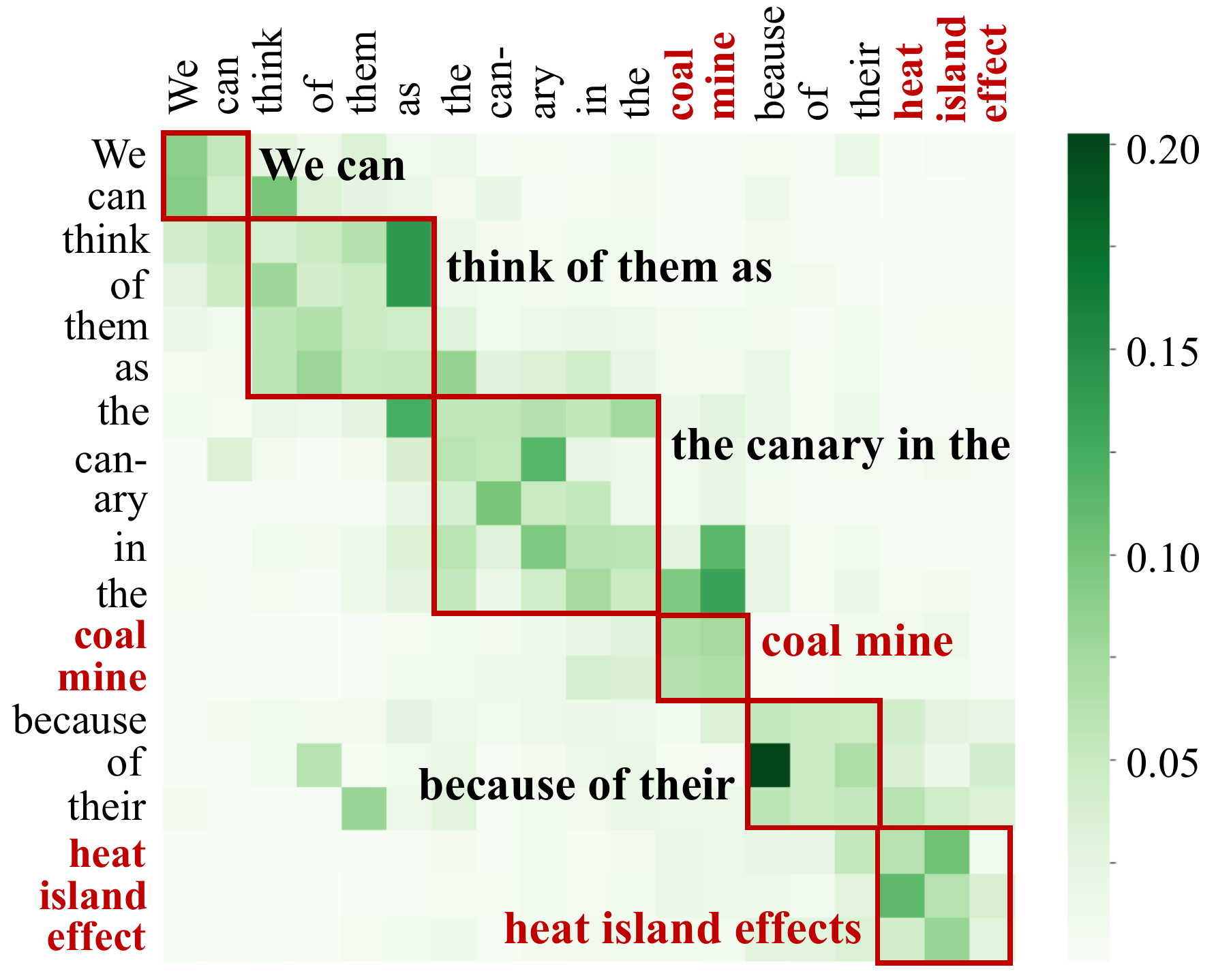}
    \caption{
    Illustration of the attention map generated by one of the pre-trained RoBERTa layers, averaged over all attention heads.
    }%
    \label{fig:attention}%
    \vspace{-2mm}
\end{figure}

\subsection{Surface-agnostic Feature Extraction}
\label{sec:attentionfeature}

To build an effective context-aware tagger for quality phrases, in addition to labels, we need to figure out the contextualized feature representation for each span.
Traditional word-identifiable features (\eg, contextualized embedding) make it easy for a classification model to overfit the silver labels by rigidly memorizing the surface names of training labels.
A degenerated name-matching model can easily achieve zero training error without really learning about any useful, generalizable features.

In principle, the recognition of a phrase should depend on the role that it plays in the sentence.
\citet{kim2019pre} show that the structure information of a sentence can be largely captured by its attention distribution.
Therefore, we propose to obtain surface-agnostic features from the attention distributions generated by a pre-trained Transformer-based language model encoder (LM), such as BERT~\cite{devlin2019bert} and RoBERTa~\cite{liu2019roberta}.

Given a sentence $[w_1, \ldots, w_N]$, we encode it with a language model pre-trained on a massive, unlabeled corpus from the same domain (\eg, general domain and scientific domain).
Suppose this language model has $L$ layers, and each layer has $H$ attention heads.
Each attention head $h$ from layer $l$ produces an attention map $\mathbf{A}^{l, h} \in \mathbb{R}^{N \times N}$ of the sentence.
The aggregated attention map from attention heads of all layers is denoted as $\mathbf{A} \in \mathbb{R}^{N \times N \times (H \cdot L)}$,
where $\mathbf{A}_{i,j} \in \mathbb{R}^{H \cdot L}$ is a vector that contains the attention scores from $w_i$ to $w_j$. 
Finally, for each candidate span $p = [w_i, \ldots, w_j]$, we denote its feature as $\mathbf{X}_{p} = \mathbf{A}_{i \ldots j, i \ldots j}$.

Ideally, the attention maps of quality phrases should reveal distinct patterns of word connections.
Figure~\ref{fig:attention} shows a real example of the generated attention map of a sentence.
The chunks on the attention map lead to a clear separation of different parts of the sentence.
From all chunks, our final span classifier (Section \ref{sec:classifier}) accurately distinguishes the quality phrases (\example{coal mine}, \example{heat island effects}) from ordinary spans (\eg, \example{We can}), indicating the informativeness of the attention features.

\noindent\textbf{Efficient Implementation.}
Thanks to the rich syntactic and semantic knowledge in the pre-trained language model, the generated attention maps are already informative enough for phrase tagging.
In this work, we adopt the RoBERTa model~\cite{liu2019roberta}, one of the state-of-the-art Transformer-based language models, as a feature extractor without the need for further fine-tuning.
We only need to apply the pre-trained RoBERTa model for one inference pass through the target corpus for feature extraction.

The overall efficiency now mainly depends on the size of the attention map, which is \tightmath{N \times N \times (H \cdot L)}.
$N$ is restricted to the length of each span during training, and for inference, we apply sentence-level encoding, with each sentence restricted to at most 64 tokens.
Depth wise, existing studies have observed considerable redundancy in the outputs of different Transformer layers, including attention distributions~\cite{gong2019efficient,gu2020transformer}.
For this reason, as the default setting of \our, we only preserve attention maps from the first $3$ layers in RoBERTa (i.e., $L = 3$).
As RoBERTa has $12$ layers in total, this saves $75\%$ of resource consumption.
We have quantitatively compared the final tagging performance of using 3 layers vs. using all $12$ layers in Section~\ref{sec:ablation}.
As the experimental results suggest, using 3 layers exhibits comparable performance with the full model.

\begin{figure}[t]
    \centering
    \includegraphics[width=0.8\linewidth]{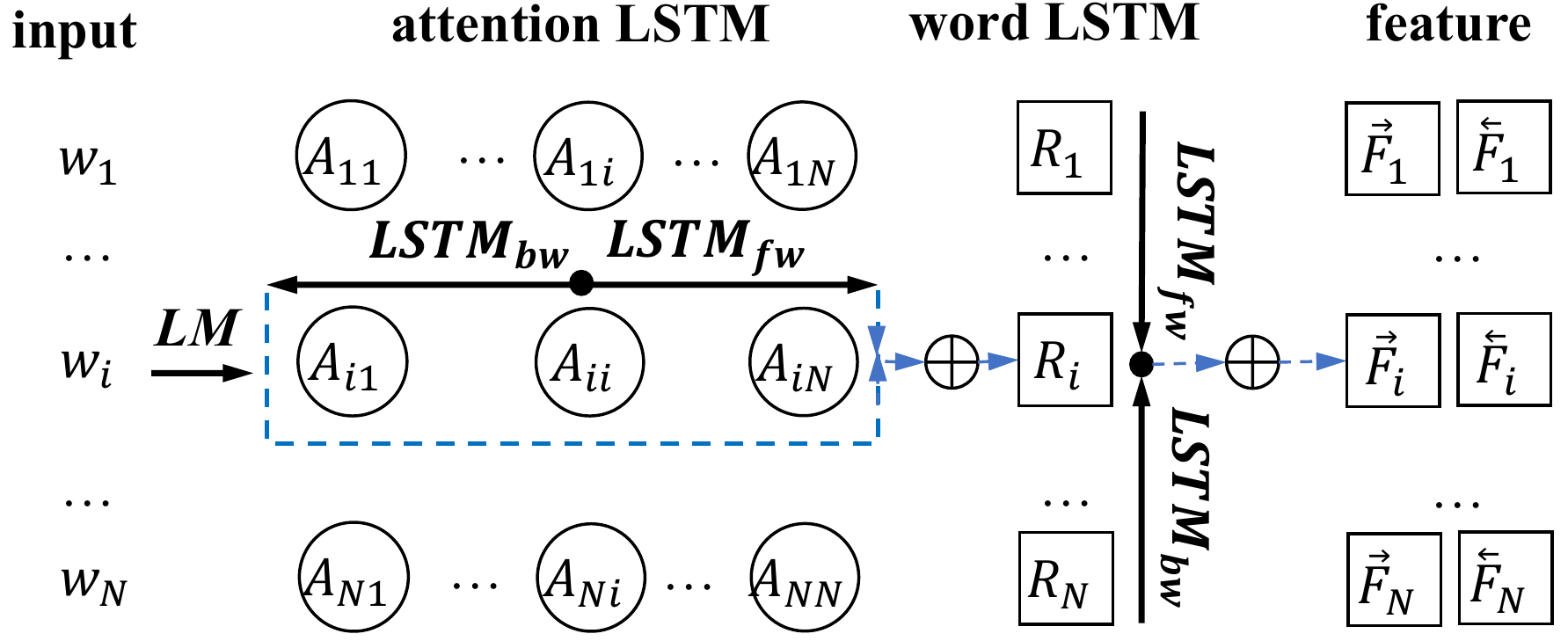}
    \caption{An alternative classifier based on attention-level LSTM.}
    \label{fig:lstm}
\end{figure}

\begin{figure*}[t]
    \centering
    \includegraphics[width=.85\linewidth]{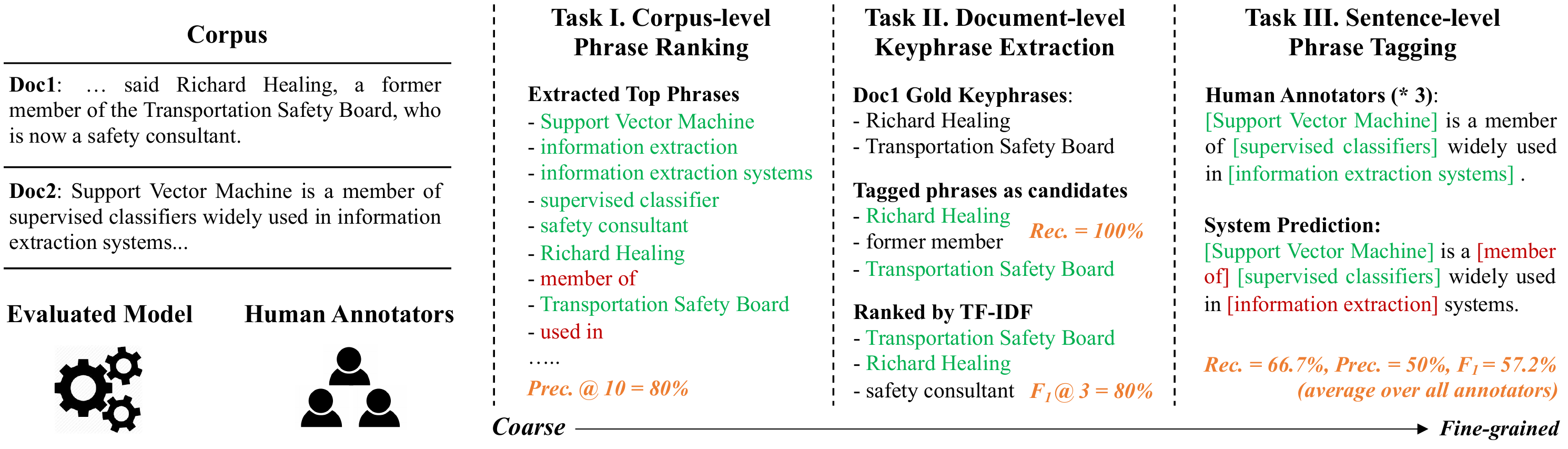}
    \caption{
    Illustration of the Three Evaluation Tasks and their Evaluation Metrics.
    }%
    \label{fig:tasks}%
    \vspace{-3mm}
\end{figure*}

\subsection{Lightweight Span Classifier}
\label{sec:classifier}

With the labels and features in-house, we are ready to build a classifier to recognize spans of quality phrases. 
Our framework is general and 
compatible with various classification models.
For the sake of efficiency, we wish to find a lightweight classifier. 

Given the attention map of a $k$-word span, an accurate classifier should effectively capture inter-word relationships from different LM layers and at different ranges.
Naturally, the attention map can be viewed as a square image of $k$ pixels for both height and width, with $H \cdot L$ channels.
We can now transform the phrase classification problem into an image classification problem: given a multi-channel image (attention map), we want to predict whether the corresponding word span is a quality phrase.
Specifically, we apply a two-layer convolutional neural network (CNN) model on the multi-channel attention map.
The output is then fed to a logistic regression layer to assign a binary label for the corresponding span. 
During the training process, the classification model $f(\cdot; \theta)$ parameterized by $\theta$ is learned by minimizing the loss over the training set $\{\mathbf{X}_{\mathcal{P}}, \mathcal{P}\}$:

$$
    \hat{\theta} = \argmin_\theta \frac{1}{|\mathcal{P}|} \sum^{|\mathcal{P}|}_{i=1} \ell(p_i, f(\mathbf{X}_{p_i}; \theta)),
$$
where $p_i \in \mathcal{P} = \bigcup\limits_{m=1}^M \{\mathcal{P}_m^-, \mathcal{P}_m^+\}$ represents the $i$-th labeled span, and $\ell$ is the binary cross entropy loss function.
The model is updated with minibatch-based stochastic gradient descent.

Note that $\theta$ here only includes the parameters in the two CNN layers and the logistic regression layer during training, which makes the training process efficient in terms of resource consumption.
In fact, the checkpoint of training parameters from each epoch can be stored in a $22$ KB file on disk.

\vpar{Alternative Classifiers.} In our study, we also considered some alternative classifiers.
One intuitive choice here is LSTM-based models following the sequence labeling framework. 
We illustrate the general idea in Figure~\ref{fig:lstm}.
Specifically, for sentence $[w_1, w_2, \ldots, w_N]$, we first encode the attention map $\mathbf{A} \in \mathbb{R}^{N \times N \times (H \cdot L)}$ with forward and backward runs of LSTM to get an attention-based word representation through the final output of both LSTMs as follows,
\begin{equation*}
    \begin{aligned}
    \overrightarrow{\mathbf{R}}_{i} &= \text{LSTM}(\mathbf{A}_{i, i}, \mathbf{A}_{i, i + 1}, \ldots, \mathbf{A}_{i, N})_{\text{last}}, \\
    \overleftarrow{\mathbf{R}}_{i} &= \text{LSTM}(\mathbf{A}_{i, i}, \mathbf{A}_{i, i - 1}, \ldots, \mathbf{A}_{i, 1})_{\text{last}}, \\
    \mathbf{R}_{i} &= [\overrightarrow{\mathbf{R}}_{i}, \overleftarrow{\mathbf{R}}_{i}], \;\; 1 \leq i \leq N.
    \end{aligned}
\end{equation*}
Another bidirectional LSTM layer is built upon the word representations $\mathbf{R}$ to extract the feature $\mathbf{F}$, \ie, 
\begin{equation*}
    \begin{aligned}
    \overrightarrow{\mathbf{F}}_{1, 2, \ldots, N} &= \text{LSTM}(\mathbf{R}_{1}, \mathbf{R}_{2}, \ldots, \mathbf{R}_{N}), \\
    \overleftarrow{\mathbf{F}}_{N, N - 1, \ldots, 1} &= \text{LSTM}(\mathbf{R}_{N}, \mathbf{R}_{N - 1}, \ldots, \mathbf{R}_{1}). \\
    \end{aligned}
\end{equation*}

\noindent Scheme wise, there are two popular labeling schemes in sequence labeling: (1) \emph{Tie-or-Break}, which is predicting whether each consecutive pair of words belong to the same phrase, and (2) \emph{BIO Tagging}, which is tagging phrases in the sentence through a Begin-Inside-Outside scheme~\cite{ramshaw1999text}. We are not using BIOES~\cite{ratinov2009design} as we focus on multi-word phrases.
For the \emph{Tie-or-Break tagging scheme}, we apply a 2-layer Multi-layer Perceptron followed by a Sigmoid classification function to predict whether the $[\overrightarrow{\mathbf{F}}_{i}, \overleftarrow{\mathbf{F}}_{i + 1}]$ representation corresponds to a tie between word $i$ and word $i + 1$ or a break. 
For the \emph{BIO tagging scheme}, in a word-wise manner, we concatenate the representations $\overrightarrow{\mathbf{F}}_{1, 2, \ldots, N}$ and $\overleftarrow{\mathbf{F}}_{N, N - 1, \ldots, 1}$ into $N$ representations for each sentence, and then send them through a Conditional Random Field (CRF) layer~\cite{lafferty2001conditional,huang2015bidirectional} to predict the BIO tags for phrases. 

Other training procedures for both of the classifiers are the same as the aforementioned default span classifier.
These alternative classifiers have comparable performance, as confirmed in Section~\ref{sec:ablation}.

\section{Experiments}

We compare our \our with previous studies on multi-word phrase mining tasks on two datasets and three tasks at different granularity: corpus-level phrase ranking, document-level keyphrase extraction, and sentence-level phrase tagging.

\subsection{Evaluation Tasks and Metrics}
\label{sec:exp-setup}
\begin{table}[]
    \renewcommand\arraystretch{.8}
    \centering
    \caption{Dataset statistics on KP20k and KPTimes. }
    \label{tab:dataset_statistics}
    \scalebox{1}{
    \small
    \begin{tabular}{l ll}
    \toprule 
    \textbf{Statistics} & \textbf{KP20k} & \textbf{KPTimes} \\
    \midrule
    & \multicolumn{2}{c}{\footnotesize{\emph{Train Set}}} \\
    \# documents & 527,090 & 259,923 \\
    \# words per document & 176 & 907 \\
    \midrule
    & \multicolumn{2}{c}{\footnotesize{\emph{Test Set}}} \\
    \# documents & 20,000 & 20,000 \\
    \# multi-word keyphrases & 37,289 & 24,920 \\
    \;\; \# unique & 24,626 & 8,970 \\
    \;\; \# absent in training corpus  & 4,171 & 2,940 \\
    \bottomrule
    \end{tabular}
    }
    \vspace{-3mm}
\end{table}

\begin{table*}[]
    \centering
    \renewcommand\tabcolsep{2pt}
    \renewcommand\arraystretch{.85}
    \caption{Evaluation results (\%) of three tasks for all compared methods on datasets on two domains.}
    \label{tab:main}
    \small
    \begin{tabular}{ll cccc cccc  cccccc}
        \toprule
        \multirow{3}{*}{\textbf{Method Type}} & \multirow{3}{*}{\textbf{Method Name}} & \multicolumn{4}{c}{\textbf{Task I: Phrase Ranking}} & \multicolumn{4}{c}{\textbf{Task \Rom{2}: KP Extract.}} &  \multicolumn{6}{c}{\textbf{Task \Rom{3}: Phrase Tagging}}  \\
        \cmidrule(lr){3-6} \cmidrule(lr){7-10} \cmidrule(lr){11-16}
        & & \multicolumn{2}{c}{\textbf{KP20k}} & \multicolumn{2}{c}{\textbf{KPTimes}} & \multicolumn{2}{c}{\textbf{KP20K}} & \multicolumn{2}{c}{\textbf{KPTimes}} & \multicolumn{3}{c}{\textbf{KP20k}} & \multicolumn{3}{c}{\textbf{KPTimes}}\\
        \cmidrule(lr){3-4} \cmidrule(lr){5-6} \cmidrule(lr){7-8} \cmidrule(lr){9-10} \cmidrule(lr){11-13} \cmidrule(lr){14-16}
        & & P\smtx{@5K} & P\smtx{@50K} & P\smtx{@5K} & P\smtx{@50K} & Rec. & F$_1$\smtx{@10} & Rec. & F$_1$\smtx{@10} & Prec. & Rec. & F$_1$  & Prec. & Rec. & F$_1$ \\
        \midrule
        \multirow{3}{*}{Pre-trained} & PKE \cite{boudin2016pke}  & -- & -- & -- & -- & 57.1 & 12.6 & 61.9 & 4.4 & 54.1 & 63.9 & 58.6 & 56.1 & 62.2 & 59.0 \\
        & Spacy \cite{spacy} & -- & -- & -- & --  & 59.5 & 15.3 & 60.8 & 8.6 &  56.3 & 68.7 & 61.9 & 61.9 & 62.9 & 62.4\\
        & StanfordNLP \cite{manning2014stanford} & -- & -- & -- & -- & 51.7 & 13.9 & 60.8 & 8.7 & 48.3 & 60.7 & 53.8 & 56.9 & 60.3 & 58.6\\
         \midrule
        \multirow{2}{*}{Distantly Supervised} & AutoPhrase \cite{shang2018automated} & 97.5 & 96.0 & 96.5 & 95.5 &  62.9 & 18.2 & 77.8 & 10.3 & 55.2 & 45.2 & 49.7 & 44.2 & 47.7 & 45.9\\
        & Wiki+RoBERTa & \textbf{100.0} & \textbf{98.5} & \textbf{99.0} & \textbf{96.5} & \textbf{73.0} & 19.2 & 64.5 & 9.4 & 58.1 & 64.2 & 61.0 & 60.9 & 65.6 & 63.2 \\
        \midrule
        \multirow{2}{*}{Unsupervised} & TopMine \cite{el2014scalable} & 81.5 & 78.0 & 85.5 & 71.0 & 53.3 & 15.0 & 63.4 & 8.5 & 39.8 & 41.4 & 40.6 & 32.0 & 36.3 & 34.0 \\
        & \our (ours) & 96.5 & 96.5 & 96.5 & 95.5 & 72.9 & \textbf{19.7} & \textbf{83.4} & \textbf{10.9} & \textbf{69.9} & \textbf{78.3} & \textbf{73.9} & \textbf{69.1} & \textbf{78.9} & \textbf{73.5} \\
        \bottomrule
    \end{tabular}
    \vspace{-3mm}
\end{table*}

We evaluate all methods on the following three tasks. Figure~\ref{fig:tasks} illustrates the tasks and evaluation metrics with some examples. 

\vpar{Task \Rom{1}: Phrase Ranking} is a popular evaluation task in previous statistics-based phrase mining work~\cite{deane2005nonparametric,li2017efficiently,el2014scalable,liu2015mining,shang2018automated}.
Specifically, it evaluates the ``global'' rank list of phrases that a method finds from the input corpus.
Since \our does not explicitly compute a ``global'' score for each phrase, we use the average logits of all occurrences of a predicted phrase to rank phrases.

In our experiments, for each method\footnote{Except for methods that does not report any form of scores for ranking.} on each dataset, we quantitatively evaluate the precision of the phrases found in the top-ranked 5,000 and 50,000 phrases, denoted as $\mathbf{P\smtx{@5K}}$ and $\mathbf{P\smtx{@50K}}$. 
Since it is expensive to hire annotators to annotate all these phrases, we estimate the precision scores by randomly sampling 200 phrases from rank lists.
Extracted phrases from different methods are shuffled and mixed before presenting to the annotators.

\vpar{Task \Rom{2}: Keyphrase Extraction} is a classic task to extract salient phrases that best summarize a document~\cite{evans1996noun}, which essentially has two stages: candidate generation and keyphrase ranking.
At the first stage, we treat all compared methods as candidate phrase extractors and evaluate the \textbf{recall} of generated candidates.
In each document, the recall measures how many gold keyphrases are extracted in the candidate list. 
For fair comparison, we preserve the same number of candidates from the rank list of each method for evaluation.

For the end-the-end performance, we apply the classic TF-IDF model to rank the candidate phrases extracted by different methods.
In each document, we follow the standard evaluation method~\cite{gallina2020large} to calculate the $F_1$ score of the top-10 ranked phrases ($\mathbf{F_1\smtx{@10}}$). 
The reported recall and $F_1$ scores are averaged in a macro way across all documents in the same dataset. 

\vpar{Task \Rom{3}: Phrase Tagging} is a fine-grained task that aims to find all occurrences of phrases in sentences.
Specifically, it evaluates the extracted phrase spans for each sentence.
We treat each phrase mining method as a sentence tagger that identifies starting and ending boundaries of phrases in a sentence. 
We randomly sample 200 sentences on each dataset and ask three annotators to tag all spans of multi-word phrases. 
Each sentence is annotated by all annotators independently, and the agreement between human annotations is around $90\%$. 
We then pool all annotations together, evaluate the predicted spans, and report the overall \textbf{precision}, \textbf{recall} and $\mathbf{F_1}$ scores.
Note that these scores are computed in a micro average fashion following previous work on entity recognition~\cite{shang2018learning}.

\subsection{Datasets}
\label{sec:dataset}

We adopt two commonly used datasets from different domains to evaluate all different methods.
\begin{itemize}[nosep,leftmargin=*]
\item \textbf{KP20k}~\cite{meng2017deep} is a collection of 
\emph{titles \& abstracts} from Computer Science papers---527,090 for training and 20,000 for testing.
\item \textbf{KPTimes}~\cite{gallina2019kptimes} consists of \emph{news articles} on New York Times from 2006 to 2017, supplemented with 10,000 more news articles from Japan Times.
In total, there are 259,923 articles for training, and 20,000 articles for testing.
\end{itemize}

\noindent 
Following~\citet{gururangan2020don}, sentence separation and tokenization is conducted with Spacy~\cite{spacy} for postprocessing.
All three tasks are evaluated on the test sets. 
Statistics of the two datasets are shown in Table~\ref{tab:dataset_statistics}.
Note that $17\%$ unique keyphrases in the test of KP20k never occur in the corresponding training corpus.
On KPTimes the absence ratio is $33\%$.
Hence, the task can be challenging for models relying on phrase frequencies and models rigidly memorizing training phrases.

\begin{table*}[]
    \renewcommand\arraystretch{.8}
    \centering
    \caption{Ablation study of \our model variants ($\%$).}
    \label{tab:ablation}
    \small
    \begin{tabular}{cccc cccc cccccc}
    \toprule
    & & & & \multicolumn{4}{c}{\textbf{KP Extract.}} &  \multicolumn{6}{c}{\textbf{Phrase Tagging}}  \\
    \cmidrule(lr){5-8} \cmidrule(lr){9-14}
    & \multicolumn{3}{c}{\textbf{Design Choices}} & \multicolumn{2}{c}{\textbf{KP20k}} & \multicolumn{2}{c}{\textbf{KPTimes}} & \multicolumn{3}{c}{\textbf{KP20k}} & \multicolumn{3}{c}{\textbf{KPTimes}} \\
    \cmidrule(lr){2-4} \cmidrule(lr){5-6} \cmidrule(lr){7-8} \cmidrule(lr){9-11} \cmidrule(lr){12-14}
    & \textbf{supervision} & \textbf{feature} & \textbf{fine-tune} & Rec. & F$_1$\smtx{@10} & Rec. & F$_1$\smtx{@10} & Prec. & Rec. & F$_1$  & Prec. & Rec. & F$_1$ \\
    \midrule
    \our & core & attention & no  & 72.9 & \textbf{19.7} & \textbf{83.4} & \textbf{10.9} & 69.9 & \textbf{78.3} & \textbf{73.9} & \textbf{69.1} & \textbf{78.9} & \textbf{73.5}  \\
    \midrule
    \multirow{4}{*}{Variants}& Wiki & attention & no & 68.7 & 17.7 & 79.4 & 10.7 & \textbf{72.1} & 71.9 & 72.0 & 64.1 & 67.6 & 65.8 \\
    & Wiki & embedding & no & 73.0 & 19.2 & 64.5 & 9.4 & 60.9 & 65.6 & 63.2 & 60.9 & 65.6 & 63.2 \\
    & core & embedding & no & 79.3 & \textbf{19.7} & 78.7 & 10.2 & 68.4 & 74.6 & 71.4 & 55.7 & 64.8 & 59.9 \\
    & core & embedding & yes & \textbf{80.3} & \textbf{19.7} & 73.9 & 9.9 & 68.6 & 74.8 & 71.6 & 53.3 & 64.5 & 59.0 \\
    \bottomrule
    \end{tabular}
    \vspace{-3mm}
\end{table*}

\subsection{Compared Methods}
\label{sec:compared-methods}
We compare the proposed method with existing methods under the same scenario, where no gold annotations for training are available.
This leads to three categories: unsupervised phrase mining methods, distantly supervised methods with an existing KB, and pre-trained off-the-shelf toolkits.
For each method that requires training (\ie, all the unsupervised and distantly supervised ones), we use the unlabeled documents from the training set for model learning.

\smallskip
\noindent For \emph{{unsupervised methods}} we consider:
\begin{itemize}[leftmargin=*,nosep]
\item \textbf{ToPMine}~\cite{el2014scalable}, the state-of-the-art unsupervised phrase mining method building upon statistical features.
\item \textbf{\our}, the proposed method in this work.
\end{itemize}

\smallskip
\noindent For \emph{distantly supervised methods}, we use silver labels generated from the \emph{Wiki Entities}, which is firstly used in~\cite{shang2018automated}.
\begin{itemize}[leftmargin=*,nosep]
    \item \textbf{AutoPhrase}~\cite{shang2018automated} leverages statistics-based phrase classifier and further enhances it with a POS-guided phrasal segmentation model for sentence tagging and phrase frequency rectification.
    \item \textbf{Wiki+RoBERTa} is a strong baseline that we propose here. 
    It can be viewed as a variant of \our with the same span prediction framework and the same pre-trained LM as our method but following distant supervision.
    Also, it uses the output states from the last layer of the pre-trained RoBERTa as feature instead of attention maps.
    As shown in~\cite{liang2020bond}, stopping the model training early is an essential intervention for distantly supervised tagging models.
    To fully unleash the potential of the Wiki+RoBERTa baseline, we \emph{manually stop} its training process \emph{after the first epoch} to avoid overfitting.
    This indeed achieves a better test performance than stopping after more epochs. 
\end{itemize}

\smallskip
\noindent For \emph{off-the-shelf toolkits} we consider the linguistic-based methods that are pre-trained with labeled pos-tagging or parsing data.
\begin{itemize}[leftmargin=*]
    \item \textbf{PKE}~\cite{boudin2016pke} is a widely used toolkit for keyphrase extraction. Its phrase mining module is a chunking model based on a supervised POS-tagging model from NLTK~\cite{bird2006nltk} and a set of grammar rules.
    \item \textbf{Spacy}~\cite{spacy} is an industrial library with a pre-trained phrase chunking model based on supervised POS tagging and parsing.
    \item \textbf{StanfordCoreNLP}~\cite{manning2014stanford} is a long recognized NLP package whose chunking model is based on dependency parsing.
\end{itemize}

\subsection{Reproduction Details}
For KPTimes, we use the official RoBERTa model pre-trained on documents from the general domain. 
On the KP20k dataset, we use the ``allenai/cs\_roberta\_base'' RoBERTa model~\cite{gururangan2020don}\myfootnote{}{https://huggingface.co/allenai/cs\_roberta\_base}.
The model is based on the standard pre-trained RoBERTa model, and then trained on unlabeled Computer Science publications.
This domain-adapted model performs slightly better on the KP20k dataset than the original model.
We adopt the Adam~\cite{kingma2015adam} optimizer with the default parameters for model training.
The learning rate is set to $0.001$.
As described in Algorithm~\ref{alg:main}, we train the classifier until its performance on the $10\%$ hold-out validation set $\mathcal{D}_{valid}$ drops.
Other details have been covered in Section~\ref{sec:method}.
We will publish our data and code base for reproduction.

\subsection{Evaluation Results}
From Table~\ref{tab:main} we can see that \our achieves the best overall performance on all the three evaluation tasks.
The performance gap becomes more vivid as the task becomes more fine-grained.

In the corpus-level phrase ranking task, most methods show very high precision (i.e., $\ge$ 95\%) on the top 50,000 mined phrases from each dataset. 
Notably, \our significantly outperforms the only other unsupervised method ToPMine and is able to perform on par with distantly supervised methods.

In the document-level keyphrase extraction task, \our has better recall than most compared methods, demonstrating a coverage of high-quality phrases.
Wiki+RoBERTa has slightly better recall on the KP20k dataset ($0.1\%$) within a reasonable range, considering Wiki+RoBERTa has access to hundreds of thousands keyphrases from Wiki Entities.
Note that \our outperforms all the compared methods on the end-to-end performance (i.e., $F_1\smtx{@10}$), which verifies its value to the application of keyphrase extraction.

In the sentence-level phrase tagging task, \our achieves $F_1$ scores of more than 73\% on both datasets, showing significant advantages (i.e., $> 10\%$ in $F_1$) over all the compared methods.
This is truly encouraging given the facts that (1) \our is an unsupervised phrase mining model that requires no human effort, and (2) even human annotators cannot fully agree with each other on some particular phrases, and have around $10\%$ disagreement on this task. 
This phrase tagging task makes clear that \our is able to find phrases much more accurately than compared methods.
In Section \ref{sec:ablation} we apply comprehensive comparison between different models on real examples, for a more straightforward visualization of the pros and cons of compared methods.

\begin{table}[]
    \renewcommand\arraystretch{.8}
    \centering
    \caption{Comparison of attention feature aggregated from different numbers of Transformer layers, evaluated on KP20k ($\%$).}
    \label{tab:layers}
    \small
    \begin{tabular}{c cc ccc}
    \toprule
    \multirow{2}{*}{\textbf{\# Layers}} & \multicolumn{2}{c}{\textbf{KP Extract.}} &  \multicolumn{3}{c}{\textbf{Phrase Tagging}}  \\
    \cmidrule(lr){2-3} \cmidrule(lr){4-6}
    & Rec. & F$_1$\smtx{@10} & Prec. & Rec. & F$_1$ \\
    \midrule
     3     & 72.9 & 19.7 &  \textbf{69.9} & \textbf{78.3} & \textbf{73.9}  \\
     12    & \textbf{81.8} & \textbf{20.6} &  69.4 & 76.8 & 72.9  \\
    \bottomrule
    \end{tabular}
    \vspace{-3mm}
\end{table}
\subsection{Ablation studies}
\label{sec:ablation}
\label{sec:ablation-supervision}
To gain deeper insights, we apply extensive ablation studies to test model variants from several aspects,
as summarized in Table~\ref{tab:ablation}.
For supervision, we compare the silver labels generated by unsupervised core phrase mining (\emph{core}), and those generated by distant supervision with Wikipedia entities (\emph{Wiki}).
For the type of features, we compare the attention map features (\emph{attention}), and the output states of RoBERTa (\emph{embedding}).

\vpar{Supervision: Core Phrase vs. Distant Supervision.}
When using the same type of feature, unsupervised models with core phrases as supervision significantly outperform distantly supervised models on most metrics by a clear gap.
The better completeness and larger volume of core phrases bring unique advantages in training context-aware tagging models, not to mention the labels are fetched from the corpus for free without relying on an external KB.
Moreover, the better diversity of core phrases effectively alleviates the risk of overfitting.
It is also worth mentioning that different from the distantly supervised embedding model, the embedding-based model trained with core phrases does not require any manual early stopping to achieve satisfying performance.

\vpar{Features: Attention vs.\ Embedding.}
When using the same type of supervision,
models with attention features are almost always better than embedding-based features. 
This verifies our intuition that word-identifiable embeddings allow the classifier to easily overfit silver labels, while the surface-agnostic attention features force the model to learn about informative contextual features, and thus having a better ability of generalization.

\vpar{Attention: First Few Layers vs.\ Full Layers.}
Table~\ref{tab:layers} compares \our trained with attention features aggregated from the first 3 layers of RoBERTa and those aggregated from all 12 layers, with intuitions explained in Section~\ref{sec:attentionfeature}.
The two models achieve comparable performance, while the small model only requires $25\%$ resource consumption.

\begin{table}[t]
    \renewcommand\arraystretch{.8}
    \renewcommand\tabcolsep{4pt}
    \centering
    \caption{Exploring LSTM-based classifiers as alternatives based on Tie-or-Break and BIO labeling schemes, evaluated on KP20k ($\%$).}
    \label{tab:classifier}
    \small
    \begin{tabular}{l cc ccc}
    \toprule
    \multirow{3}{*}{\textbf{Classifier}} & \multicolumn{2}{c}{\textbf{KP Extract.}} &  \multicolumn{3}{c}{\textbf{Phrase Tagging}}  \\
    \cmidrule(lr){2-3} \cmidrule(lr){4-6}
    & Rec. & F$_1$\smtx{@10} & Prec. & Rec. & F$_1$ \\
    \midrule
    CNN (default in \our) & 68.1 & 18.7  & 69.9 & \textbf{78.3} & \textbf{73.9} \\
    LSTM w/ Tie-or-Break     & \textbf{72.4} & \textbf{19.3} & 68.1 & 72.3 & 70.1 \\
    LSTM w/ BIO  & 66.2 & 18.1 & \textbf{71.0} & 76.7 & 73.7       \\
    \bottomrule
    \end{tabular}
    \vspace{-3mm}
\end{table}

\vpar{Alternative Classifiers.}
Table~\ref{tab:classifier} compares our model with the the Tie-or-Break classifier and the BIO classifier as introduced in Section \ref{sec:classifier}. 
Overall, the alternative  classifiers have comparable performances, indicating the ability of our proposed method to generalize to different tagging schemes and model architectures.

\begin{table*}[t]
    \footnotesize
    \renewcommand\arraystretch{1.3}
        \caption{Sentences tagged with different methods described in Section \ref{sec:compared-methods}.}
        \vspace{-1mm}
        \label{tb:case_clean}
        \scalebox{.94}{
        \begin{tabular}{@{}r@{\hskip 1em}p{.93\columnwidth}@{\hskip 2em}p{1.05\columnwidth}@{}}
            \toprule
            & \multicolumn{1}{c}{\textbf{KP20k}} & \multicolumn{1}{c}{\textbf{KPTimes}}\\
            \midrule
            
            \textbf{Spacy} 
            & 
            We are interested in improving the {Varshamov bound} for \watag{finite values} of length $n$ and \watag{minimum distance} $d$. We employ a \watag{counting lemma} to this end which we find particularly useful in relation to \watag{Varshamov graphs} .
            &
            The \watag{United States} , at least theoretically , taxes companies on their \watag{global profits} . 
            But companies with a lot of \watag{intellectual property} -- notably \watag{technology and pharmaceutical companies} -- get away with paying a fraction of that amount .
            \\
            \textbf{AutoPhrase}
            &
            We are interested in improving the \cetag{Varshamov bound} for {finite values} of length $n$ and \cetag{minimum distance} $d$. We employ a \cetag{counting lemma} to this end which we find particularly useful in relation to {Varshamov graphs} .
            &
            The \cetag{United States} , at least theoretically , {taxes companies} on their {global profits} . 
            {But companies} with a lot of \cetag{intellectual property} -- notably \cetag{technology and pharmaceutical companies} -- get away with paying a fraction of that amount .
            \\
            \textbf{RoBERTa}
            &
            We are interested in improving the {Varshamov bound} for {finite values} of length $n$ and {minimum distance} $d$. We employ a \wetag{counting lemma} to this end which we find particularly useful in relation to {Varshamov graphs} .
            &
            The \wetag{United States} , at least theoretically , \wetag{taxes companies} on their \wetag{global profits} .
            \wetag{But companies} with a lot of \wetag{intellectual property} -- notably technology and \wetag{pharmaceutical companies} -- get away with paying a fraction of that amount .
            \\
            \textbf{\our}
            &
            We are interested in improving the \catag{Varshamov bound} for \catag{finite values} of length $n$ and \catag{minimum distance} $d$. We employ a \catag{counting lemma} to this end which we find particularly useful in relation to \catag{Varshamov graphs} .
            &
            The \catag{United States} , at least theoretically , taxes companies on their \catag{global profits} .
            But companies with a lot of \catag{intellectual property} -- notably technology and \catag{pharmaceutical companies} -- get away with paying a fraction of that amount .
            \\

            \bottomrule
        \end{tabular}
        }
    \vspace{-3mm}
\end{table*}
\subsection{Case Studies}
\label{sec:case_study}

In spite of the reasonably high quality of the silver labels, we are curious about whether our final span classifier is robust to the noisy silver labels.
To this end, we feed the silver labels to the span classifier and investigate the predicted probability scores.
Table~\ref{tab:noise} presents the silver labels with probabilities below $1\%$ and above $99\%$ respectively.
As it shows, our classifier successfully distinguishes high-quality core phrases from noisy spans, including typos (\emph{italic font}) that happen to be used consistently in some document.
The classifier draws a clear line between these two kinds of spans based on their attention features, which reflect their distinct roles in sentences.
We have attempted to remove the low-score ones from the silver labels and re-train the classifier, however, the final performance changes little.
This further verifies the robustness of our model, and its ability to capture general context features rather than rigid memorization.

Table~\ref{tb:case_clean} presents sentences tagged with representative methods from each category.
As it shows, pre-trained 
models like Spacy can hardly adapt to a new domain without human annotations.
For instance, it fails to recognize \example{Varshamov bound} as a phrase for recognizing \example{bound} as a verb.
Statistics-based methods like AutoPhrase tend to miss uncommon phrases in the corpus, such as \example{Varshamov graphs}, \example{finite values}, and \example{global profiles}.
The widely used distantly supervised methods based on word representations from a pre-trained language model (\eg, RoBERTa) can easily overfit the phrases in the KB, even though we have applied manual early stopping.
The consequence of rigid memorization comes in two folds.
First, the model can miss a lot of out-of-KB phrases, such as the terminologies in KP20k.
Second, it can recognize false phrases just because they have similar surface names with real phrases.
In the example from KPTimes, the model recognizes \example{taxes companies} and \example{but companies} as two phrases, while \example{taxes} is used as a verb in this sentence, and \example{but} is a conjunction word.
Overall, the results generated by \our are more accurate. 
There is also an interesting case in the example from KPTimes, where RoBERTa and \our recognize \example{pharmaceutical companies} as a complete phrase, while Spacy and AutoPhrase think \example{technology and} is also part of the phrase.
It is debatable which one is better: both results can contribute to a high-quality phrase vocabulary.
In fact, even human annotators cannot achieve perfect agreement in their independent annotations.
Dynamically adjusting the granularity of tagged phrases according to different end tasks remains a valuable research problem for further studies.

\newcommand{\typo}[1]{\emph{#1}} 
\newcommand{\prob}[1]{\textcolor{gblue}{ }} 

\begin{table}[]
    \renewcommand\arraystretch{.1}
    \centering
    \small
    \caption{
    Examples from silver training labels with extremely high and low quality scores $f(\cdot : \theta)$ estimated by \our.
    The results show that \our is robust to noises in training labels.
    }
    \label{tab:noise}
    \scalebox{.86}{
    \begin{tabular}{@{}p{.5\columnwidth}p{.5\columnwidth}@{}}
    \toprule
    \multicolumn{2}{c}{\textbf{KP20k}}\\
    \multicolumn{1}{c}{${f(\cdot : \theta) > 99\%}$} & \multicolumn{1}{c}{${f(\cdot : \theta) < 1\%}$}\\
    \cmidrule(lr){1-1} \cmidrule(lr){2-2}
    model identification, data structures, release dates, VLSI design, product development, network flow, finite precision,
    watermark detection, model selection, path planning, network security, data centers, source code,
    \ldots
    &
    times fewer\prob{.000005}, prescriptions implies\prob{.000009}, algorithms require estimating\prob{.00002}, significantly improves performance\prob{.00002},
    including discontinuities\prob{.00004}, significantly reduce power consumption, factors include\prob{.00004}, \typo{considered byTitterington} \ldots\\
    \midrule 
    \multicolumn{2}{c}{\textbf{KPTimes}} \\
    \multicolumn{1}{c}{${f(\cdot : \theta) > 99\%}$} & \multicolumn{1}{c}{${f(\cdot : \theta) < 1\%}$}\\
    \cmidrule(lr){1-1} \cmidrule(lr){2-2}
    Davis Cup, Ivy League, no-fly zone, Tour Championship, tax returns, City Hall, home runs, detention center, operating system, Ryder Cup, space stations, ice packs, White House,
    Jersey City, board games, tax cuts,
    \ldots
    &
    \typo{PThe percentage}, 11th title, departments began telling officers, category includes workers, attacks including,
    74th career win, including political, countries including Spain, including banking, including mobile, \ldots\\
    \bottomrule
    
    \end{tabular}
    }
    \vspace{-3.5mm}
\end{table}

\section{Related Work}
Phrase mining is a long studied task~\cite{frantzi2000automatic,deane2005nonparametric,el2014scalable,liu2015mining,shang2018automated}. 
Due to the broad applicability of phrases to text-associated tasks, supervision signals would be expensive to obtain for vast domains. 
Unsupervised approaches have been proposed to extract phrases from many different angles, most importantly, language grammar~\cite{neubig2011unsupervised, spacy, manning2014stanford} and text statistics~\cite{el2014scalable}. 
Our work utilizes contextualized features from Transformer-based language models~\cite{devlin2019bert, liu2019roberta}, therefore, lifts the unnecessary requirement of frequency in statistics-based methods and alleviates requirements of expert-crafted grammar rules. 
Through experiments of three different tasks (i.e., corpus-level phrase ranking, document-level keyphrase extraction, and sentence-level phrase tagging), our method shows great performance improvement over previous methods. 

Another line of research studies on distant supervision signals, such as existing knowledge bases~\cite{shang2018automated,wang2020mining}. 
They typically use knowledge base entries (\eg, Wiki Entities from~\cite{shang2018automated}) to string-match a corpus to obtain supervision signals in their first step. Such matching does not take into account how n-grams exist in the corpus, and as we show, could lead to partial matching of phrases, thus bringing bias to the phrase mining tool trained (\eg, \example{heat island effect} is usually matched into \example{island effect}). 
Our core phrase mining method, while being unsupervised, looks into the context of each n-gram to find max patterns and is able to find more complete phrases that serve as a better supervision signal to \our contextualized feature based classifier.

We use attention maps from pre-trained Transformer-based language models to identify phrases since they carry inter-relation information of tokens~\cite{clark2019does, kim2019pre}. \citet{clark2019does} showed that a sufficient amount of linguistic knowledge, such as noun determiners and objects of verbs and prepositions, are captured by attention maps of BERT. 
Moreover, using only attention maps, one can train a model to perform dependency parsing~\cite{clark2019does} and constituency tree construction~\cite{kim2019pre} relatively well. 
Our work utilizes this powerful nature of attention maps and treats them as the only feature to identify quality phrases. 
Furthermore, through comparing with the output states of RoBERTa, we show that using attention is less likely to overfit and has a more robust generalization.

\section{Conclusions}
We explore phrase tagging in an unsupervised and context-aware manner. 
Our proposed method, \our, shows clear improvement on performance for three quality-measuring tasks on two datasets in different domains. 
Further experimental studies reveal the strength of our two major components: our unsupervised core phrase mining finds more diverse, complete phrases in context than string-matching from some knowledge bases; 
our use of attention features unleashes the rich linguistic knowledge contained in pre-trained neural language models. 
By leveraging surface-agnostic context features, our model removes the frequency requirement in statistics-based models and alleviates the overfitting issue in embedding-based models.

We plan to explore the following directions in future studies. 
First, our study shows that the combination of silver labels and attention is robust and contains sufficient linguistic knowledge. 
This idea of unsupervised learning is worth exploring in other text mining tasks, such as coreference resolution~\cite{mccarthy1995using}, dependency parsing~\cite{kubler2009dependency}, and named entity recognition~\cite{nadeau2007survey}. 
Second, the imperfection of distant supervision calls for a more effective way to incorporate large-scale unlabeled corpus with existing knowledge bases for more accurate prediction and more intelligent reasoning.
\begin{acks}
Research was supported in part by US DARPA KAIROS Program No. FA8750-19-2-1004, SocialSim Program No. W911NF-17-C-0099, National Science Foundation IIS-19-56151, IIS-17-41317, and IIS 17-04532, NSF Convergence Accelerator under award OIA-2040727, and the Molecule Maker Lab Institute: An AI Research Institutes program supported by NSF under Award No. 2019897. 
\end{acks}

\bibliographystyle{ACM-Reference-Format}
\bibliography{sample-base}


\begin{thebibliography}{37}


\ifx \showCODEN    \undefined \def \showCODEN     #1{\unskip}     \fi
\ifx \showDOI      \undefined \def \showDOI       #1{#1}\fi
\ifx \showISBNx    \undefined \def \showISBNx     #1{\unskip}     \fi
\ifx \showISBNxiii \undefined \def \showISBNxiii  #1{\unskip}     \fi
\ifx \showISSN     \undefined \def \showISSN      #1{\unskip}     \fi
\ifx \showLCCN     \undefined \def \showLCCN      #1{\unskip}     \fi
\ifx \shownote     \undefined \def \shownote      #1{#1}          \fi
\ifx \showarticletitle \undefined \def \showarticletitle #1{#1}   \fi
\ifx \showURL      \undefined \def \showURL       {\relax}        \fi
\providecommand\bibfield[2]{#2}
\providecommand\bibinfo[2]{#2}
\providecommand\natexlab[1]{#1}
\providecommand\showeprint[2][]{arXiv:#2}

\bibitem[\protect\citeauthoryear{Alt{\i}nel and Ganiz}{Alt{\i}nel and
  Ganiz}{2018}]%
        {altinel2018semantic}
\bibfield{author}{\bibinfo{person}{Berna Alt{\i}nel} {and}
  \bibinfo{person}{Murat~Can Ganiz}.} \bibinfo{year}{2018}\natexlab{}.
\newblock \showarticletitle{Semantic text classification: A survey of past and
  recent advances}.
\newblock \bibinfo{journal}{\emph{Information Processing \& Management}}
  \bibinfo{volume}{54}, \bibinfo{number}{6} (\bibinfo{year}{2018}).
\newblock


\bibitem[\protect\citeauthoryear{Bird}{Bird}{2006}]%
        {bird2006nltk}
\bibfield{author}{\bibinfo{person}{Steven Bird}.}
  \bibinfo{year}{2006}\natexlab{}.
\newblock \showarticletitle{NLTK: the natural language toolkit}. In
  \bibinfo{booktitle}{\emph{Proceedings of the COLING/ACL 2006 Interactive
  Presentation Sessions}}. \bibinfo{pages}{69--72}.
\newblock


\bibitem[\protect\citeauthoryear{Boudin}{Boudin}{2016}]%
        {boudin2016pke}
\bibfield{author}{\bibinfo{person}{Florian Boudin}.}
  \bibinfo{year}{2016}\natexlab{}.
\newblock \showarticletitle{PKE: an open source python-based keyphrase
  extraction toolkit}. In \bibinfo{booktitle}{\emph{Proceedings of COLING 2016,
  the 26th International Conference on Computational Linguistics: System
  Demonstrations}}. \bibinfo{pages}{69--73}.
\newblock


\bibitem[\protect\citeauthoryear{Clark, Khandelwal, Levy, and Manning}{Clark
  et~al\mbox{.}}{2019}]%
        {clark2019does}
\bibfield{author}{\bibinfo{person}{Kevin Clark}, \bibinfo{person}{Urvashi
  Khandelwal}, \bibinfo{person}{Omer Levy}, {and}
  \bibinfo{person}{Christopher~D Manning}.} \bibinfo{year}{2019}\natexlab{}.
\newblock \showarticletitle{What Does BERT Look at? An Analysis of BERT’s
  Attention}. In \bibinfo{booktitle}{\emph{Proceedings of the 2019 ACL Workshop
  BlackboxNLP: Analyzing and Interpreting Neural Networks for NLP}}.
  \bibinfo{pages}{276--286}.
\newblock


\bibitem[\protect\citeauthoryear{Croft, Turtle, and Lewis}{Croft
  et~al\mbox{.}}{1991}]%
        {croft1991use}
\bibfield{author}{\bibinfo{person}{W~Bruce Croft}, \bibinfo{person}{Howard~R
  Turtle}, {and} \bibinfo{person}{David~D Lewis}.}
  \bibinfo{year}{1991}\natexlab{}.
\newblock \showarticletitle{The use of phrases and structured queries in
  information retrieval}. In \bibinfo{booktitle}{\emph{Proceedings of the 14th
  annual international ACM SIGIR conference on Research and development in
  information retrieval}}. \bibinfo{pages}{32--45}.
\newblock


\bibitem[\protect\citeauthoryear{Deane}{Deane}{2005}]%
        {deane2005nonparametric}
\bibfield{author}{\bibinfo{person}{Paul Deane}.}
  \bibinfo{year}{2005}\natexlab{}.
\newblock \showarticletitle{A nonparametric method for extraction of candidate
  phrasal terms}. In \bibinfo{booktitle}{\emph{Proceedings of the 43rd Annual
  Meeting of the Association for Computational Linguistics (ACL’05)}}.
  \bibinfo{pages}{605--613}.
\newblock


\bibitem[\protect\citeauthoryear{Devlin, Chang, Lee, and Toutanova}{Devlin
  et~al\mbox{.}}{2019}]%
        {devlin2019bert}
\bibfield{author}{\bibinfo{person}{Jacob Devlin}, \bibinfo{person}{Ming-Wei
  Chang}, \bibinfo{person}{Kenton Lee}, {and} \bibinfo{person}{Kristina
  Toutanova}.} \bibinfo{year}{2019}\natexlab{}.
\newblock \showarticletitle{BERT: Pre-training of Deep Bidirectional
  Transformers for Language Understanding}. In
  \bibinfo{booktitle}{\emph{Proceedings of the 2019 Conference of the North
  American Chapter of the Association for Computational Linguistics: Human
  Language Technologies, Volume 1 (Long and Short Papers)}}.
  \bibinfo{pages}{4171--4186}.
\newblock


\bibitem[\protect\citeauthoryear{El-Kishky, Song, Wang, Voss, and
  Han}{El-Kishky et~al\mbox{.}}{2014}]%
        {el2014scalable}
\bibfield{author}{\bibinfo{person}{Ahmed El-Kishky}, \bibinfo{person}{Yanglei
  Song}, \bibinfo{person}{Chi Wang}, \bibinfo{person}{Clare~R Voss}, {and}
  \bibinfo{person}{Jiawei Han}.} \bibinfo{year}{2014}\natexlab{}.
\newblock \showarticletitle{Scalable Topical Phrase Mining from Text Corpora}.
\newblock \bibinfo{journal}{\emph{Proceedings of the VLDB Endowment}}
  \bibinfo{volume}{8}, \bibinfo{number}{3} (\bibinfo{year}{2014}).
\newblock


\bibitem[\protect\citeauthoryear{Evans and Zhai}{Evans and Zhai}{1996}]%
        {evans1996noun}
\bibfield{author}{\bibinfo{person}{David~A Evans} {and}
  \bibinfo{person}{Chengxiang Zhai}.} \bibinfo{year}{1996}\natexlab{}.
\newblock \showarticletitle{Noun-phrase analysis in unrestricted text for
  information retrieval}. In \bibinfo{booktitle}{\emph{Proceedings of the 34th
  annual meeting on Association for Computational Linguistics}}.
  \bibinfo{pages}{17--24}.
\newblock


\bibitem[\protect\citeauthoryear{Finch}{Finch}{2016}]%
        {finch2016linguistic}
\bibfield{author}{\bibinfo{person}{Geoffrey Finch}.}
  \bibinfo{year}{2016}\natexlab{}.
\newblock \bibinfo{booktitle}{\emph{Linguistic terms and concepts}}.
\newblock \bibinfo{publisher}{Macmillan International Higher Education}.
\newblock


\bibitem[\protect\citeauthoryear{Frantzi, Ananiadou, and Mima}{Frantzi
  et~al\mbox{.}}{2000}]%
        {frantzi2000automatic}
\bibfield{author}{\bibinfo{person}{Katerina Frantzi}, \bibinfo{person}{Sophia
  Ananiadou}, {and} \bibinfo{person}{Hideki Mima}.}
  \bibinfo{year}{2000}\natexlab{}.
\newblock \showarticletitle{Automatic recognition of multi-word terms:. the
  c-value/nc-value method}.
\newblock \bibinfo{journal}{\emph{International journal on digital libraries}}
  \bibinfo{volume}{3}, \bibinfo{number}{2} (\bibinfo{year}{2000}),
  \bibinfo{pages}{115--130}.
\newblock


\bibitem[\protect\citeauthoryear{Gallina, Boudin, and Daille}{Gallina
  et~al\mbox{.}}{2019}]%
        {gallina2019kptimes}
\bibfield{author}{\bibinfo{person}{Ygor Gallina}, \bibinfo{person}{Florian
  Boudin}, {and} \bibinfo{person}{B{\'e}atrice Daille}.}
  \bibinfo{year}{2019}\natexlab{}.
\newblock \showarticletitle{KPTimes: A Large-Scale Dataset for Keyphrase
  Generation on News Documents}. In \bibinfo{booktitle}{\emph{Proceedings of
  the 12th International Conference on Natural Language Generation}}.
  \bibinfo{pages}{130--135}.
\newblock


\bibitem[\protect\citeauthoryear{Gallina, Boudin, and Daille}{Gallina
  et~al\mbox{.}}{2020}]%
        {gallina2020large}
\bibfield{author}{\bibinfo{person}{Ygor Gallina}, \bibinfo{person}{Florian
  Boudin}, {and} \bibinfo{person}{B{\'e}atrice Daille}.}
  \bibinfo{year}{2020}\natexlab{}.
\newblock \showarticletitle{Large-scale evaluation of keyphrase extraction
  models}. In \bibinfo{booktitle}{\emph{Proceedings of the ACM/IEEE Joint
  Conference on Digital Libraries in 2020}}. \bibinfo{pages}{271--278}.
\newblock


\bibitem[\protect\citeauthoryear{Gong, He, Li, Qin, Wang, and Liu}{Gong
  et~al\mbox{.}}{2019}]%
        {gong2019efficient}
\bibfield{author}{\bibinfo{person}{Linyuan Gong}, \bibinfo{person}{Di He},
  \bibinfo{person}{Zhuohan Li}, \bibinfo{person}{Tao Qin},
  \bibinfo{person}{Liwei Wang}, {and} \bibinfo{person}{Tieyan Liu}.}
  \bibinfo{year}{2019}\natexlab{}.
\newblock \showarticletitle{Efficient training of bert by progressively
  stacking}. In \bibinfo{booktitle}{\emph{International Conference on Machine
  Learning}}. PMLR, \bibinfo{pages}{2337--2346}.
\newblock


\bibitem[\protect\citeauthoryear{Gu, Liu, Yu, Li, Chen, and Han}{Gu
  et~al\mbox{.}}{2020}]%
        {gu2020transformer}
\bibfield{author}{\bibinfo{person}{Xiaotao Gu}, \bibinfo{person}{Liyuan Liu},
  \bibinfo{person}{Hongkun Yu}, \bibinfo{person}{Jing Li},
  \bibinfo{person}{Chen Chen}, {and} \bibinfo{person}{Jiawei Han}.}
  \bibinfo{year}{2020}\natexlab{}.
\newblock \showarticletitle{On the Transformer Growth for Progressive BERT
  Training}.
\newblock \bibinfo{journal}{\emph{arXiv preprint arXiv:2010.12562}}
  (\bibinfo{year}{2020}).
\newblock


\bibitem[\protect\citeauthoryear{Gururangan, Marasovi{\'c}, Swayamdipta, Lo,
  Beltagy, Downey, and Smith}{Gururangan et~al\mbox{.}}{2020}]%
        {gururangan2020don}
\bibfield{author}{\bibinfo{person}{Suchin Gururangan}, \bibinfo{person}{Ana
  Marasovi{\'c}}, \bibinfo{person}{Swabha Swayamdipta}, \bibinfo{person}{Kyle
  Lo}, \bibinfo{person}{Iz Beltagy}, \bibinfo{person}{Doug Downey}, {and}
  \bibinfo{person}{Noah~A Smith}.} \bibinfo{year}{2020}\natexlab{}.
\newblock \showarticletitle{Don’t Stop Pretraining: Adapt Language Models to
  Domains and Tasks}. In \bibinfo{booktitle}{\emph{Proceedings of the 58th
  Annual Meeting of the Association for Computational Linguistics}}.
  \bibinfo{pages}{8342--8360}.
\newblock


\bibitem[\protect\citeauthoryear{Honnibal, Montani, Van~Landeghem, and
  Boyd}{Honnibal et~al\mbox{.}}{2020}]%
        {spacy}
\bibfield{author}{\bibinfo{person}{Matthew Honnibal}, \bibinfo{person}{Ines
  Montani}, \bibinfo{person}{Sofie Van~Landeghem}, {and}
  \bibinfo{person}{Adriane Boyd}.} \bibinfo{year}{2020}\natexlab{}.
\newblock \bibinfo{booktitle}{\emph{{spaCy: Industrial-strength Natural
  Language Processing in Python}}}.
\newblock


\bibitem[\protect\citeauthoryear{Huang, Xu, and Yu}{Huang
  et~al\mbox{.}}{2015}]%
        {huang2015bidirectional}
\bibfield{author}{\bibinfo{person}{Zhiheng Huang}, \bibinfo{person}{Wei Xu},
  {and} \bibinfo{person}{Kai Yu}.} \bibinfo{year}{2015}\natexlab{}.
\newblock \showarticletitle{Bidirectional LSTM-CRF models for sequence
  tagging}.
\newblock \bibinfo{journal}{\emph{arXiv preprint arXiv:1508.01991}}
  (\bibinfo{year}{2015}).
\newblock


\bibitem[\protect\citeauthoryear{Kim, Choi, Edmiston, and Lee}{Kim
  et~al\mbox{.}}{2019}]%
        {kim2019pre}
\bibfield{author}{\bibinfo{person}{Taeuk Kim}, \bibinfo{person}{Jihun Choi},
  \bibinfo{person}{Daniel Edmiston}, {and} \bibinfo{person}{Sang-goo Lee}.}
  \bibinfo{year}{2019}\natexlab{}.
\newblock \showarticletitle{Are Pre-trained Language Models Aware of Phrases?
  Simple but Strong Baselines for Grammar Induction}. In
  \bibinfo{booktitle}{\emph{International Conference on Learning
  Representations}}.
\newblock


\bibitem[\protect\citeauthoryear{Kingma and Ba}{Kingma and Ba}{2015}]%
        {kingma2015adam}
\bibfield{author}{\bibinfo{person}{Diederik~P Kingma} {and}
  \bibinfo{person}{Jimmy Ba}.} \bibinfo{year}{2015}\natexlab{}.
\newblock \showarticletitle{Adam: A Method for Stochastic Optimization}. In
  \bibinfo{booktitle}{\emph{ICLR (Poster)}}.
\newblock


\bibitem[\protect\citeauthoryear{K{\"u}bler, McDonald, and Nivre}{K{\"u}bler
  et~al\mbox{.}}{2009}]%
        {kubler2009dependency}
\bibfield{author}{\bibinfo{person}{Sandra K{\"u}bler}, \bibinfo{person}{Ryan
  McDonald}, {and} \bibinfo{person}{Joakim Nivre}.}
  \bibinfo{year}{2009}\natexlab{}.
\newblock \showarticletitle{Dependency parsing}.
\newblock \bibinfo{journal}{\emph{Synthesis lectures on human language
  technologies}} \bibinfo{volume}{1}, \bibinfo{number}{1}
  (\bibinfo{year}{2009}), \bibinfo{pages}{1--127}.
\newblock


\bibitem[\protect\citeauthoryear{Lafferty, McCallum, and Pereira}{Lafferty
  et~al\mbox{.}}{2001}]%
        {lafferty2001conditional}
\bibfield{author}{\bibinfo{person}{John Lafferty}, \bibinfo{person}{Andrew
  McCallum}, {and} \bibinfo{person}{Fernando~CN Pereira}.}
  \bibinfo{year}{2001}\natexlab{}.
\newblock \showarticletitle{Conditional random fields: Probabilistic models for
  segmenting and labeling sequence data}.
\newblock  (\bibinfo{year}{2001}).
\newblock


\bibitem[\protect\citeauthoryear{Li, Yang, Wang, and Cui}{Li
  et~al\mbox{.}}{2017}]%
        {li2017efficiently}
\bibfield{author}{\bibinfo{person}{Bing Li}, \bibinfo{person}{Xiaochun Yang},
  \bibinfo{person}{Bin Wang}, {and} \bibinfo{person}{Wei Cui}.}
  \bibinfo{year}{2017}\natexlab{}.
\newblock \showarticletitle{Efficiently Mining High Quality Phrases from
  Texts}. In \bibinfo{booktitle}{\emph{AAAI}}.
\newblock


\bibitem[\protect\citeauthoryear{Liang, Yu, Jiang, Er, Wang, Zhao, and
  Zhang}{Liang et~al\mbox{.}}{2020}]%
        {liang2020bond}
\bibfield{author}{\bibinfo{person}{Chen Liang}, \bibinfo{person}{Yue Yu},
  \bibinfo{person}{Haoming Jiang}, \bibinfo{person}{Siawpeng Er},
  \bibinfo{person}{Ruijia Wang}, \bibinfo{person}{Tuo Zhao}, {and}
  \bibinfo{person}{Chao Zhang}.} \bibinfo{year}{2020}\natexlab{}.
\newblock \showarticletitle{Bond: Bert-assisted open-domain named entity
  recognition with distant supervision}. In
  \bibinfo{booktitle}{\emph{Proceedings of the 26th ACM SIGKDD International
  Conference on Knowledge Discovery \& Data Mining}}.
  \bibinfo{pages}{1054--1064}.
\newblock


\bibitem[\protect\citeauthoryear{Liu, Shang, Wang, Ren, and Han}{Liu
  et~al\mbox{.}}{2015}]%
        {liu2015mining}
\bibfield{author}{\bibinfo{person}{Jialu Liu}, \bibinfo{person}{Jingbo Shang},
  \bibinfo{person}{Chi Wang}, \bibinfo{person}{Xiang Ren}, {and}
  \bibinfo{person}{Jiawei Han}.} \bibinfo{year}{2015}\natexlab{}.
\newblock \showarticletitle{Mining quality phrases from massive text corpora}.
  In \bibinfo{booktitle}{\emph{Proceedings of the 2015 ACM SIGMOD International
  Conference on Management of Data}}. \bibinfo{pages}{1729--1744}.
\newblock


\bibitem[\protect\citeauthoryear{Liu, Ott, Goyal, Du, Joshi, Chen, Levy, Lewis,
  Zettlemoyer, and Stoyanov}{Liu et~al\mbox{.}}{2019}]%
        {liu2019roberta}
\bibfield{author}{\bibinfo{person}{Yinhan Liu}, \bibinfo{person}{Myle Ott},
  \bibinfo{person}{Naman Goyal}, \bibinfo{person}{Jingfei Du},
  \bibinfo{person}{Mandar Joshi}, \bibinfo{person}{Danqi Chen},
  \bibinfo{person}{Omer Levy}, \bibinfo{person}{Mike Lewis},
  \bibinfo{person}{Luke Zettlemoyer}, {and} \bibinfo{person}{Veselin
  Stoyanov}.} \bibinfo{year}{2019}\natexlab{}.
\newblock \showarticletitle{Roberta: A robustly optimized bert pretraining
  approach}.
\newblock  (\bibinfo{year}{2019}).
\newblock


\bibitem[\protect\citeauthoryear{Manning, Surdeanu, Bauer, Finkel, Bethard, and
  McClosky}{Manning et~al\mbox{.}}{2014}]%
        {manning2014stanford}
\bibfield{author}{\bibinfo{person}{Christopher~D Manning},
  \bibinfo{person}{Mihai Surdeanu}, \bibinfo{person}{John Bauer},
  \bibinfo{person}{Jenny~Rose Finkel}, \bibinfo{person}{Steven Bethard}, {and}
  \bibinfo{person}{David McClosky}.} \bibinfo{year}{2014}\natexlab{}.
\newblock \showarticletitle{The Stanford CoreNLP natural language processing
  toolkit}. In \bibinfo{booktitle}{\emph{Proceedings of 52nd annual meeting of
  the association for computational linguistics: system demonstrations}}.
  \bibinfo{pages}{55--60}.
\newblock


\bibitem[\protect\citeauthoryear{McCarthy and Lehnert}{McCarthy and
  Lehnert}{1995}]%
        {mccarthy1995using}
\bibfield{author}{\bibinfo{person}{Joseph~F McCarthy} {and}
  \bibinfo{person}{Wendy~G Lehnert}.} \bibinfo{year}{1995}\natexlab{}.
\newblock \showarticletitle{Using decision trees for conference resolution}. In
  \bibinfo{booktitle}{\emph{Proceedings of the 14th international joint
  conference on Artificial intelligence-Volume 2}}.
  \bibinfo{pages}{1050--1055}.
\newblock


\bibitem[\protect\citeauthoryear{Meng, Zhao, Han, He, Brusilovsky, and
  Chi}{Meng et~al\mbox{.}}{2017}]%
        {meng2017deep}
\bibfield{author}{\bibinfo{person}{Rui Meng}, \bibinfo{person}{Sanqiang Zhao},
  \bibinfo{person}{Shuguang Han}, \bibinfo{person}{Daqing He},
  \bibinfo{person}{Peter Brusilovsky}, {and} \bibinfo{person}{Yu Chi}.}
  \bibinfo{year}{2017}\natexlab{}.
\newblock \showarticletitle{Deep Keyphrase Generation}. In
  \bibinfo{booktitle}{\emph{Proceedings of the 55th Annual Meeting of the
  Association for Computational Linguistics (Volume 1: Long Papers)}}.
  \bibinfo{pages}{582--592}.
\newblock


\bibitem[\protect\citeauthoryear{Nadeau and Sekine}{Nadeau and Sekine}{2007}]%
        {nadeau2007survey}
\bibfield{author}{\bibinfo{person}{David Nadeau} {and} \bibinfo{person}{Satoshi
  Sekine}.} \bibinfo{year}{2007}\natexlab{}.
\newblock \showarticletitle{A survey of named entity recognition and
  classification}.
\newblock \bibinfo{journal}{\emph{Lingvisticae Investigationes}}
  \bibinfo{volume}{30}, \bibinfo{number}{1} (\bibinfo{year}{2007}),
  \bibinfo{pages}{3--26}.
\newblock


\bibitem[\protect\citeauthoryear{Neubig, Watanabe, Sumita, Mori, and
  Kawahara}{Neubig et~al\mbox{.}}{2011}]%
        {neubig2011unsupervised}
\bibfield{author}{\bibinfo{person}{Graham Neubig}, \bibinfo{person}{Taro
  Watanabe}, \bibinfo{person}{Eiichiro Sumita}, \bibinfo{person}{Shinsuke
  Mori}, {and} \bibinfo{person}{Tatsuya Kawahara}.}
  \bibinfo{year}{2011}\natexlab{}.
\newblock \showarticletitle{An unsupervised model for joint phrase alignment
  and extraction}. In \bibinfo{booktitle}{\emph{Proceedings of the 49th Annual
  Meeting of the Association for Computational Linguistics: Human Language
  Technologies}}. \bibinfo{pages}{632--641}.
\newblock


\bibitem[\protect\citeauthoryear{Ramshaw and Marcus}{Ramshaw and
  Marcus}{1999}]%
        {ramshaw1999text}
\bibfield{author}{\bibinfo{person}{Lance~A Ramshaw} {and}
  \bibinfo{person}{Mitchell~P Marcus}.} \bibinfo{year}{1999}\natexlab{}.
\newblock \showarticletitle{Text chunking using transformation-based learning}.
\newblock In \bibinfo{booktitle}{\emph{Natural language processing using very
  large corpora}}. \bibinfo{publisher}{Springer}, \bibinfo{pages}{157--176}.
\newblock


\bibitem[\protect\citeauthoryear{Ratinov and Roth}{Ratinov and Roth}{2009}]%
        {ratinov2009design}
\bibfield{author}{\bibinfo{person}{Lev Ratinov} {and} \bibinfo{person}{Dan
  Roth}.} \bibinfo{year}{2009}\natexlab{}.
\newblock \showarticletitle{Design challenges and misconceptions in named
  entity recognition}. In \bibinfo{booktitle}{\emph{Proceedings of the
  Thirteenth Conference on Computational Natural Language Learning
  (CoNLL-2009)}}. \bibinfo{pages}{147--155}.
\newblock


\bibitem[\protect\citeauthoryear{Shang, Liu, Jiang, Ren, Voss, and Han}{Shang
  et~al\mbox{.}}{2018b}]%
        {shang2018automated}
\bibfield{author}{\bibinfo{person}{Jingbo Shang}, \bibinfo{person}{Jialu Liu},
  \bibinfo{person}{Meng Jiang}, \bibinfo{person}{Xiang Ren},
  \bibinfo{person}{Clare~R Voss}, {and} \bibinfo{person}{Jiawei Han}.}
  \bibinfo{year}{2018}\natexlab{b}.
\newblock \showarticletitle{Automated phrase mining from massive text corpora}.
\newblock \bibinfo{journal}{\emph{IEEE Transactions on Knowledge and Data
  Engineering}} \bibinfo{volume}{30}, \bibinfo{number}{10}
  (\bibinfo{year}{2018}), \bibinfo{pages}{1825--1837}.
\newblock


\bibitem[\protect\citeauthoryear{Shang, Liu, Gu, Ren, Ren, and Han}{Shang
  et~al\mbox{.}}{2018a}]%
        {shang2018learning}
\bibfield{author}{\bibinfo{person}{Jingbo Shang}, \bibinfo{person}{Liyuan Liu},
  \bibinfo{person}{Xiaotao Gu}, \bibinfo{person}{Xiang Ren},
  \bibinfo{person}{Teng Ren}, {and} \bibinfo{person}{Jiawei Han}.}
  \bibinfo{year}{2018}\natexlab{a}.
\newblock \showarticletitle{Learning Named Entity Tagger using Domain-Specific
  Dictionary}. In \bibinfo{booktitle}{\emph{Proceedings of the 2018 Conference
  on Empirical Methods in Natural Language Processing}}.
\newblock


\bibitem[\protect\citeauthoryear{Wang, Zhu, Jiang, Zhang, Wang, Ni, Xie, and
  Xiao}{Wang et~al\mbox{.}}{2020}]%
        {wang2020mining}
\bibfield{author}{\bibinfo{person}{Li Wang}, \bibinfo{person}{Wei Zhu},
  \bibinfo{person}{Sihang Jiang}, \bibinfo{person}{Sheng Zhang},
  \bibinfo{person}{Keqiang Wang}, \bibinfo{person}{Yuan Ni},
  \bibinfo{person}{Guotong Xie}, {and} \bibinfo{person}{Yanghua Xiao}.}
  \bibinfo{year}{2020}\natexlab{}.
\newblock \showarticletitle{Mining Infrequent High-Quality Phrases from
  Domain-Specific Corpora}. In \bibinfo{booktitle}{\emph{Proceedings of the
  29th ACM International Conference on Information \& Knowledge Management}}.
  \bibinfo{pages}{1535--1544}.
\newblock


\bibitem[\protect\citeauthoryear{Williams, Lessard, Desu, Clark, Bagrow,
  Danforth, and Dodds}{Williams et~al\mbox{.}}{2015}]%
        {williams2015zipf}
\bibfield{author}{\bibinfo{person}{Jake~Ryland Williams},
  \bibinfo{person}{Paul~R Lessard}, \bibinfo{person}{Suma Desu},
  \bibinfo{person}{Eric~M Clark}, \bibinfo{person}{James~P Bagrow},
  \bibinfo{person}{Christopher~M Danforth}, {and}
  \bibinfo{person}{Peter~Sheridan Dodds}.} \bibinfo{year}{2015}\natexlab{}.
\newblock \showarticletitle{Zipf’s law holds for phrases, not words}.
\newblock \bibinfo{journal}{\emph{Scientific reports}} \bibinfo{volume}{5},
  \bibinfo{number}{1} (\bibinfo{year}{2015}), \bibinfo{pages}{1--7}.
\newblock


\end{thebibliography}

\appendix

\end{document}